\documentclass[conference]{IEEEtran}
\usepackage{graphicx}
\usepackage{amsmath}
\usepackage{multirow}
\usepackage{booktabs}
\usepackage{makecell} 
\usepackage{tabularx}
\usepackage[numbers]{natbib}
\usepackage{multicol}
\usepackage[bookmarks=true]{hyperref}
\usepackage{cuted}
\usepackage{caption}
\usepackage{algorithm}
\usepackage{algorithmic}
\usepackage{xcolor}
\usepackage{tcolorbox}
\usepackage{bm}
\usepackage{amssymb}

\begin{document}

\title{Learning to Evolve: Multi-modal Interactive Fields for Robust Humanoid Navigation in Dynamic Environments}



\author{ Peifeng Jiang\textsuperscript{1}, Hong Liu\textsuperscript{1,*}, Jin Jin\textsuperscript{2}, Wenshuai Wang\textsuperscript{1}, Xia Li\textsuperscript{3} \\
  \textsuperscript{1} State Key Laboratory of General Artificial Intelligence,  Peking University, Shenzhen Graduate School \\
  \textsuperscript{2} Oxford Robotics Institute, University of Oxford \\
  \textsuperscript{3} Institute for Machine Learning, Department of Computer Science, ETH Zurich \\
  \texttt{jpf@stu.pku.edu.cn, hongliu@pku.edu.cn, jinjin@robots.ox.ac.uk,} \\ \texttt{wangws@stu.pku.edu.cn, xia.li@inf.ethz.ch}}


\maketitle

\begin{strip}
    \centering
    \includegraphics[width=0.95\textwidth]{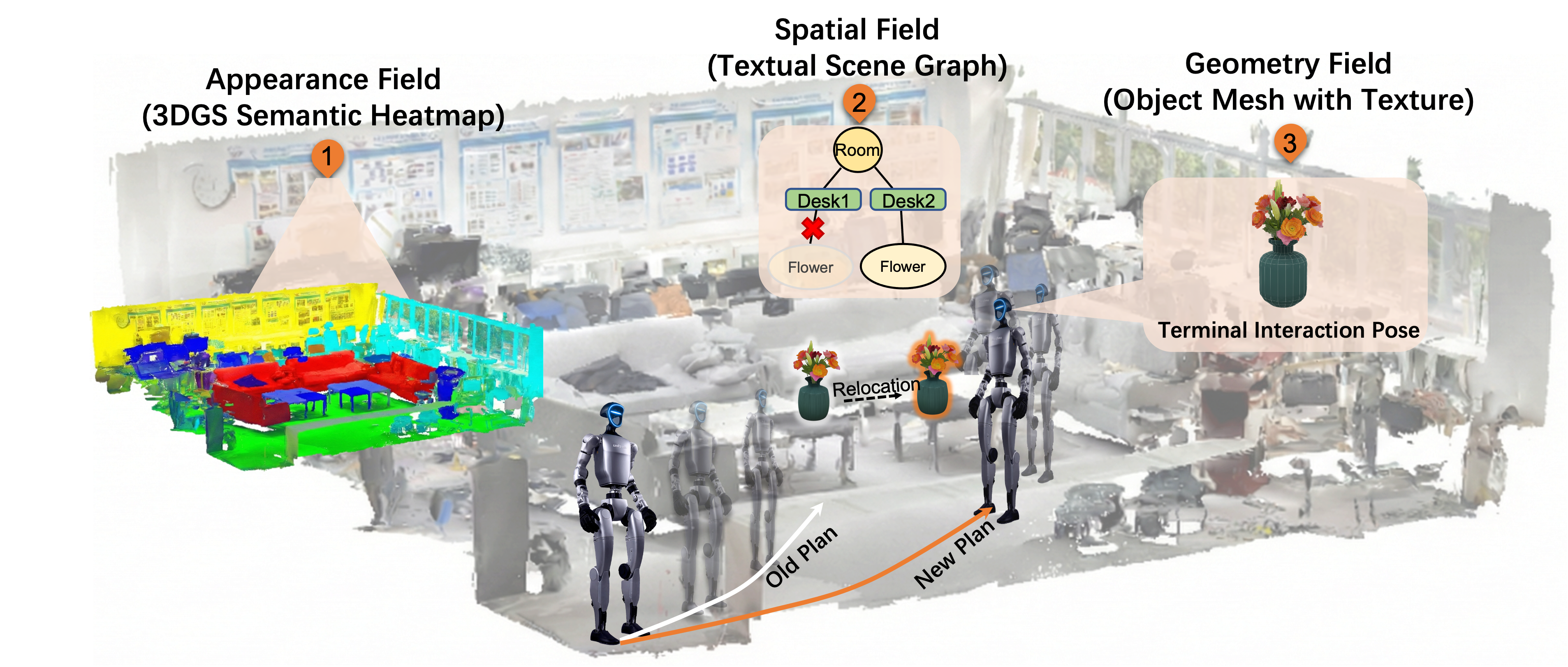}
    \captionof{figure}{\textbf{Multi-modal Interactive Fields (MIF) for Robust Humanoid Navigation.} 
We propose a hierarchical framework composed of three coupled fields: \textbf{(1) Appearance Field:} Provides dense semantic grounding for robust view synthesis. 
\textbf{(2) Spatial Field:} Maintains a dynamic topological Scene Graph that updates upon object relocation, enabling seamless re-planning from obsolete paths to correct trajectories. 
\textbf{(3) Geometry Field:} Generates water-tight meshes to ensure Interaction Pose Safety (IPS) before manipulation.}
    \label{fig:teaser}
\end{strip}

\begin{abstract}
Safe manipulation-oriented navigation for humanoid robots requires scene memory that remains reliable under locomotion-induced perceptual distortion, environmental changes, and interaction-level geometric safety constraints.
Existing semantic mapping and scene-graph systems are difficult to deploy directly in this setting because they often assume stable camera trajectories, static environments, or coarse object geometry.
We introduce the \textbf{Multi-modal Interactive Field (MIF)}, a humanoid-oriented system that integrates confidence-aware semantic 3D Gaussian Splatting, discrepancy-triggered spatial memory updates, and task-driven geometric reconstruction within a closed-loop perception-adaptation pipeline.
MIF couples three fields: an uncertainty-aware 3DGS Appearance Field that suppresses gait-induced blur, a Spatial Field that maintains topological memory, and a Geometry Field that supports \textbf{Interaction Pose Safety (IPS)} before manipulation.
A discrepancy detection score is introduced to separate locomotion-induced false-positive changes from persistent changes and updates only locally inconsistent regions.
On a \textbf{Unitree-G1 humanoid} in a real dynamic office, MIF improves relocation success in non-static environments from 12\% to \textbf{94\%} compared with static scene-graph memory, while reducing semantic memory footprint by 91.4\% through feature distillation for practical online operation.
Project page and code: \url{https://ziya-jiang.github.io/MIF-homepage/}.
\end{abstract}

\begin{keywords}
Semantic Scene Understanding, Safe Navigation, Humanoid Robots, Active Perception, 3D Gaussian Splatting, Multiplex Interactive Learning
\end{keywords}

\IEEEpeerreviewmaketitle

\section{Introduction}
Humanoid robots are increasingly expected to operate in human-centric environments where navigation is tightly coupled with subsequent physical interaction.
Approximate navigation to an object is therefore insufficient: the robot must reach a terminal stance and body configuration from which manipulation is collision-free, kinematically reachable, and stable.
This requires scene memory that grounds open-vocabulary semantics into reliable geometry while remaining robust to locomotion-induced perceptual distortion and environmental changes.
Unlike wheeled platforms, humanoids introduce bipedal dynamics, including torso oscillations and abrupt viewpoint shifts, that can degrade egocentric RGB-D observations during mapping~\cite{stasse2006real, kajita20013d}.
This challenges a common assumption behind many semantic mapping and scene-graph pipelines, such as HOV-SG~\cite{Werby-RSS-24} and ConceptGraphs~\cite{gu2024conceptgraphs}, which are typically built from stable, quasi-static observations.
At the same time, such systems often follow a map-once-query-forever paradigm, whereas human environments contain relocated, removed, or newly introduced objects.
Together, these assumptions expose two coupled failure modes for manipulation-oriented humanoid navigation:
\begin{itemize}
    \item \textbf{P1: Locomotion-Induced Semantic-Geometric Distortion. }
    Humanoid locomotion induces camera jitter and noisy state estimation for pose that can cause unstable observations to be fused into the map.
    This produces blurred semantics and noisy geometry, degrading both object grounding and the surface evidence needed to compute a safe terminal interaction pose.    
    \item \textbf{P2: Map-Reality Mismatch under Scene Changes.} 
    Real-world layouts change as objects are relocated, removed, or added, while static scene memories may continue to guide the robot toward obsolete coordinates.
    For a humanoid, stale memory not merely cause error in semantic grounding: it can lead to unsafe interaction poses, unnecessary re-planning, or failures to recognize task-relevant changes.
\end{itemize}

To address these coupled failures, we propose the Multi-modal Interactive Field (MIF), a humanoid-oriented system that maintains appearance, spatial, and geometry fields within a closed-loop perception-adaptation pipeline, as shown in Fig.~\ref{fig:teaser}.

For P1, the Appearance Field builds a confidence-aware semantic 3D Gaussian Splatting representation in which each Gaussian carries a reliability estimate for identifying gait-corrupted primitives.
This reliability gate suppresses gait-corrupted primitives during rendering and scene-graph construction, reducing the propagation of locomotion artifacts into semantic grounding.
For interaction-level safety, the Geometry Field selects better views around the target object and uses a Flow Matching-based model~\cite{chen2025sam, lipman2022flow} to recover object-centric meshes on demand.
The recovered mesh provides the geometric proxy required by  {Interaction Pose Safety (IPS)}, allowing navigation completion to depend on collision-awareness, kinematically reachability, and stability rather than just coordinate-level arrival.

For P2, MIF uses a closed-loop  {Interaction and Adaptation Mechanism} that treats scene memory as locally revisable during task execution.
Rather than globally re-scanning the environment, the system monitors a multi-modal discrepancy score $\mathcal{D}$ between the live local graph and the stored global graph.
Because $\mathcal{D}$ is gated by reliability and aggregates positional, semantic, and relational evidence, it separates locomotion-induced false-positive changes from persistent scene changes.
When $\mathcal{D}$ exceeds the selected threshold, MIF triggers a spatially selective update that revises only the inconsistent region.
This local evolution loop avoids the cost of global re-mapping while supporting adaptation to relocated, removed, and newly added objects.
Accordingly, our contributions emphasize the integrated system:

\begin{itemize}
    \item MIF provides a humanoid-oriented integration of confidence-aware semantic 3D Gaussian Splatting, VLM-based scene-graph reasoning, and object-centric geometry recovery for manipulation-safety-aware navigation.
    
    \item A discrepancy-triggered local update mechanism compares live and stored graph evidence, filters locomotion-induced false positives through confidence gating, and revises only inconsistent regions of the spatial memory.
    
    \item Interaction Pose Safety (IPS) is formulated as a system-level navigation criterion, supported by target-centered view selection and on-demand Flow Matching-based mesh recovery.
    
    \item Real-world validation on a Unitree G1 humanoid in a dynamic office reports task success, memory, and latency evidence for online deployment. 
    Under non-static environment settings, MIF improves relocation success from 12\% to 94\% compared with static scene-graph memory.
\end{itemize}

\section{RELATED WORK}
\label{sec:related_work}

\subsection{Semantic 3D Scene Representations}
Neural Radiance Fields (NeRF)~\cite{mildenhall2021nerf} and Instant NGP~\cite{muller2022instant} have advanced continuous 3D scene modeling.
3D Gaussian Splatting (3DGS)~\cite{kerbl20233d} further offers explicit primitives and real-time rendering, making it attractive for robotic mapping and online scene representations.
To attach open-vocabulary semantics to such representations, prior work~\cite{kerr2023lerf, kirillov2023segment, peng2023openscene, shi2024language, kobayashi2022decomposing, engelmann2024opennerf, gadre2023cows} integrates vision foundation models~\cite{kirillov2023segment, radford2021learning, oquab2024dinov2} into 3D fields.
Extensions such as Gaussian Grouping~\cite{gaussian_grouping}, LangSplat~\cite{qin2024langsplat}, and Feature Splatting~\cite{qiu2024featuresplatting} distill or attach high-dimensional semantic features to Gaussian primitives for object-centric understanding, language querying, and editing.
LangSplat reduces language-feature storage through a scene-wise autoencoder, while LatentBKI~\cite{wilson2025latentbki} studies open-vocabulary mapping in visual-language latent spaces with quantifiable uncertainty.
These methods reduce the cost of open-vocabulary 3D querying, but dense semantic embeddings can still impose nontrivial memory and latency overhead for online humanoid deployment.

In parallel, motion-deblurring 3DGS methods such as Deblur-GS~\cite{oh2024deblurgs} reconstruct sharper Gaussian radiance fields from camera motion-blurred images, while uncertainty-aware methods such as PUP 3D-GS~\cite{hanson2025pup} estimate Gaussian sensitivity for pruning and efficient rendering.
However, these lines of work are usually studied as reconstruction, feature storage, or pruning problems, rather than as a coupled humanoid system that must use the reliability signal for image rendering, scene-graph construction, and manipulation-safety verification.
For humanoid deployment, bipedal locomotion can introduce gait-induced artifacts, while dense semantic features can also create memory and latency costs.

MIF addresses these constraints by integrating confidence-gated semantic 3DGS with feature distillation: the former suppresses gait-corrupted primitives for P1, while the latter is an efficiency-enabling design for practical online deployment rather than a standalone novelty claim.

\subsection{3D Scene Graphs and Hierarchical Mapping}
3D Scene Graphs (3DSGs) provide a structured abstraction by organizing environments into floors, rooms, objects, and their relations~\cite{armeni20193d, Werby-RSS-24, wu2021scenegraphfusion, wald2020learning, bavle2022situational}.
Early systems such as Kimera-Semantics~\cite{rosinol20203d} focused on geometric-semantic consistency, while Hydra~\cite{hughes2022hydra} introduced real-time spatial perception.
With the advent of large language and vision-language models~\cite{hurst2024gpt}, recent frameworks~\cite{Werby-RSS-24, rana2023sayplan, chalvatzaki2023learning, ni2024grid, yin2024sg} leverage symbolic or language-conditioned reasoning for complex tasks.

Dynamic scene-graph and spatio-temporal mapping methods, such as Khronos~\cite{schmid2024khronos}, further consider how scene memories evolve when objects move, disappear, or appear.
ConceptGraphs~\cite{gu2024conceptgraphs} can also be adapted with periodic rescanning, providing a useful rescan-style baseline for testing whether scene memories can be refreshed after changes.
Nevertheless, static scene-graph memories expose a map-once-query-forever failure mode, while dynamic or rescan-style alternatives may rely on revisiting, periodic rescanning, or sensing assumptions that differ from our non-revisiting humanoid navigation protocol.
Moreover, graph nodes and coarse object boxes often lack the surface-level geometry needed to evaluate manipulation-ready terminal poses.

MIF bridges this gap by coupling the Spatial Field with an on-demand Geometry Field, while using discrepancy-triggered updates to revise only inconsistent regions of the memory during navigation.

\subsection{Generative Geometry for Embodied AI}
Robotic interaction requires reliable geometric priors, especially when only partial observations are available.
Learning-based mesh generation has been studied through methods such as BSP-based reconstruction, Scan2Mesh, AtlasNet-style surface generation, and PolyGen~\cite{chen2020bsp, dai2019scan2mesh, groueix2018papier, nash2020polygen}.
Generative models, including diffusion-based~\cite{ho2020denoising} and Flow Matching-based~\cite{lipman2022flow} approaches, have further improved shape recovery from sparse or partial observations.
Recent studies such as MeshGPT~\cite{siddiqui2024meshgpt} and SAM-3D~\cite{chen2025sam} explore object-centric completion that can support downstream manipulation.
However, these generative priors are typically used as offline or object-centric modules, rather than being integrated with online navigation and memory adaptation.

MIF incorporates a Flow Matching-based Geometry Field that recovers object-centric meshes \textit{on demand} during the approaching phase and uses them as geometric proxies for Interaction Pose Safety (IPS) verification.
Our use of generative geometry is therefore not a standalone mesh-generation contribution; rather, it provides an on-demand geometric proxy selected from target-centered views and used by IPS during humanoid navigation.

\subsection{Active Perception and Interactive Learning}
Interactive perception~\cite{bohg2017interactive} studies how an agent's actions can reduce perceptual ambiguity.
Related work spans low-level active SLAM~\cite{bajcsy1988active, zhao2022perception, xu2021invariant, kim2013perception, chaplot2020learning} and high-level semantic exploration~\cite{yokoyama2024vlfm, zhu2025move, chen2023not, chaplot2020object}.
Modern foundation models provide useful zero-shot priors, but they can still hallucinate or produce stale semantic assumptions in novel environments~\cite{chang2024goat, chen2023not}.
Although active perception connects sensing with action, humanoid-specific constraints such as locomotion-induced viewpoint oscillations remain less explored.
In contrast to coverage-oriented exploration, MIF uses two task-driven mechanisms: target-centered view selection for interaction geometry and discrepancy-triggered local memory revision for map-reality conflicts.
This makes scene memory locally revisable during navigation, moving the humanoid beyond passive use of a static map.

\begin{figure*}[t]
    \centering
    \includegraphics[width=0.95\textwidth]{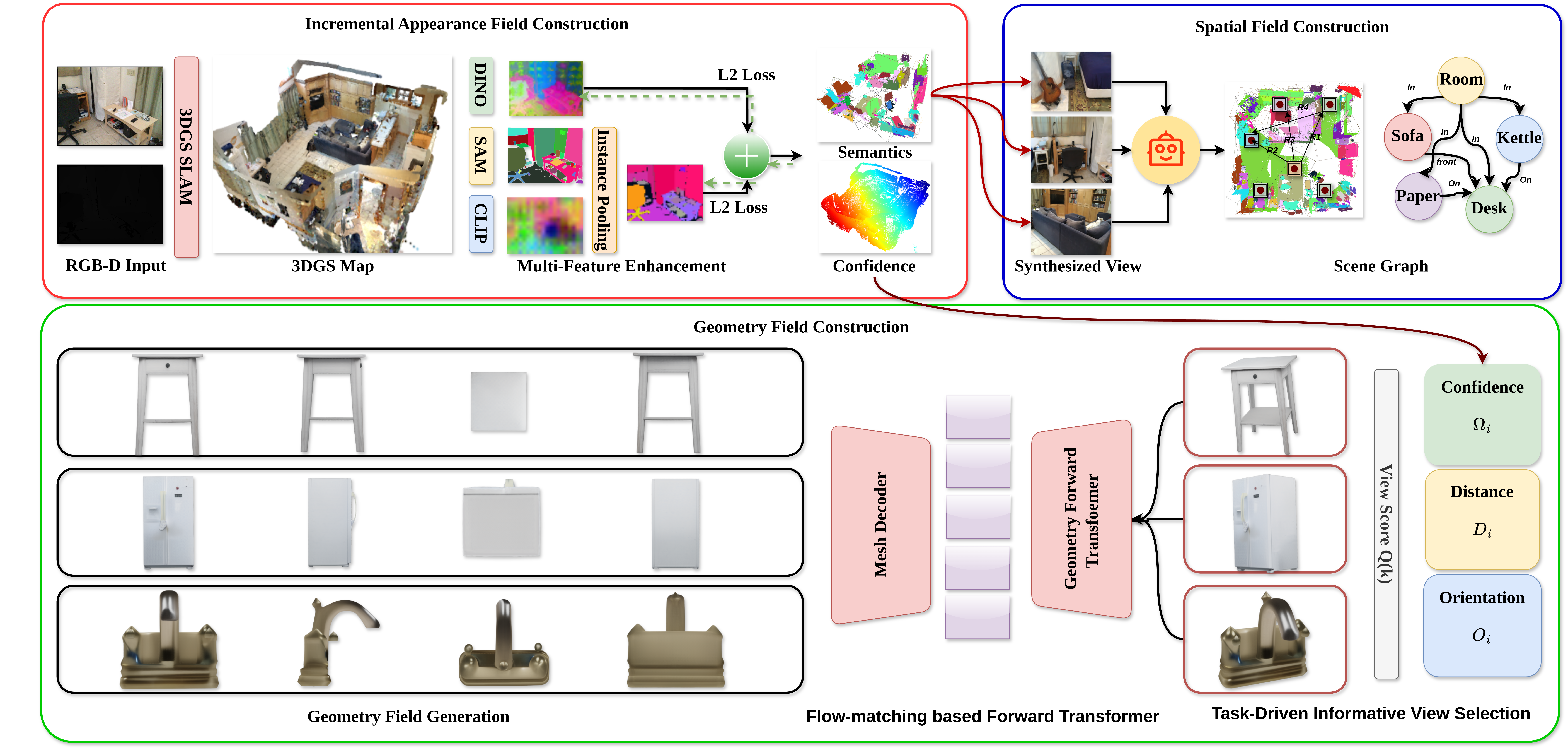}
    \caption{\textbf{Overview of the Multi-modal Interactive Fields (MIF) framework.} \textbf{(Red)} Incremental Appearance Field: Constructs a dense semantic base by fusing 3DGS SLAM~\cite{yan2023gs} with multi-modal features~\cite{oquab2024dinov2, radford2021learning} to generate confidence-aware maps. \textbf{(Blue)} Spatial Field: Abstracts the dense map into a topological Scene Graph, utilizing VLMs~\cite{hurst2024gpt} to reason over object relationships from synthesized views. \textbf{(Green)} Geometry Field: Recovers high-fidelity meshes for interaction. It employs a Task-Driven View Selection strategy for object-centric geometry recovery by adapting a Flow-matching Transformer~\cite{chen2025sam}.}
    \label{fig:overview}
\end{figure*}

\section{METHODOLOGY}
\label{sec:methodology}
We propose the  {Multi-modal Interactive Field (MIF)}, a humanoid-oriented system representation $\mathcal{M} = \{ \mathcal{F}_{app}, \mathcal{F}_{spat}, \mathcal{F}_{geom} \}$ for safe manipulation-oriented navigation from egocentric RGB-D observations.
Unlike static mapping pipelines, MIF is designed to mitigate \textit{locomotion-induced distortion} (P1) and \textit{map-reality mismatches} (P2) by grounding language queries into spatial memory and interaction-ready geometry.
The pipeline operates as a persistent perception-adaptation loop that incrementally stabilizes appearance, updates spatial memory, and verifies interaction geometry.
The overview of the Multi-modal Interactive Fields (MIF) framework is shown in Fig~\ref{fig:overview}. In the following sections, we detail:
(i) the construction of the confidence-gated  {Appearance Field} $\mathcal{F}_{app}$ via jitter-aware semantic 3DGS (Sec.~\ref{sec:appearance_field});
(ii) the abstraction of the  {Spatial Field} $\mathcal{F}_{spat}$ from stabilized rendered views for semantic grounding and mismatch detection (Sec.~\ref{sec:spatial_field}); and
(iii) the on-demand generation of the  {Geometry Field} $\mathcal{F}_{geom}$ for  {Interaction Pose Safety} (IPS) verification (Sec.~\ref{sec:geometry_field}).


\subsection{Appearance Field Construction ($\mathcal{F}_{app}$)}
\label{sec:appearance_field}
The Appearance Field $\mathcal{F}_{app}$ serves as the sensory grounding layer of MIF.
Unlike standard reconstruction pipelines that assume relatively stable camera motion, $\mathcal{F}_{app}$ is designed to mitigate the semantic-geometric distortion (P1) caused by bipedal locomotion.
Rather than directly fusing all raw frames, it uses per-Gaussian reliability to downweight observations likely affected by high-frequency torso oscillations and local pose uncertainty, producing a more view-consistent semantic radiance field.

We represent the field as a collection of 3D Gaussian primitives $\mathcal{G} = \{G_i\}$.
Each primitive is augmented with a confidence scalar $C_i \in [0, 1]$, which acts as a reliability gate for identifying and suppressing gait-corrupted primitives.

\subsubsection{Jitter-Aware Confidence Estimation}
To reduce transient artifacts such as motion blur and pose-induced misalignment, we introduce a differentiable confidence module.
The key observation is that gait-corrupted primitives tend to remain optimization-unstable across training, whereas stable high-texture boundaries usually converge after multi-view optimization.
Accordingly, we compute $C_i$ from each Gaussian's weighted gradient magnitude and opacity, normalized by their global means as $g_{n,i}$ and $\alpha_{n,i}$, respectively.
The subscript $n$ denotes normalization by the corresponding global mean.
The gradient term aggregates instability from position, scale/rotation, color, and opacity updates, while opacity reflects whether a primitive has accumulated stable structural evidence.

The confidence $C_i$ is then computed through a smooth gating mechanism that penalizes high instability while preserving structurally supported primitives:
\begin{equation}
    C_i = \exp(-\beta g_{n,i}) + (1 - \exp(-\beta g_{n,i})) \cdot \sigma(\gamma \alpha_{n,i}(1 - g_{n,i})) \label{eq:confidence},
\end{equation}
where $\sigma(\cdot)$ denotes the sigmoid function, which maps the term $\gamma \alpha_{n,i}(1-g_{n,i})$ into the range $[0,1]$. $\beta$ controls the penalty on optimization instability and $\gamma$ controls the opacity-based reliability scaling.
This mechanism downweights low-confidence, ghosting-like artifacts in rendering and object reliability estimation, reducing their influence on the subsequent spatial graph.

\subsubsection{Open-Vocabulary Feature Distillation}
To enable online semantic querying with bounded memory, we distill high-dimensional CLIP/DINOv2 features~\cite{radford2021learning, oquab2024dinov2} into a compact 32-dimensional latent space $\mathbf{f}_i^{distill} \in \mathbb{R}^{32}$.
This compression reduces the memory footprint needed to maintain large-scale semantic maps and supports practical online deployment (see Sec.~\ref{sec:exp_performance}).

A co-trained MLP decoder projects the latent features back to the original CLIP/DINOv2 feature space for open-vocabulary matching.
Feature rendering follows standard alpha-blending but is weighted by the confidence metric so that unstable primitives contribute less to semantic evidence:
\begin{equation}
    \mathbf{F}^{distill}(\mathbf{p}) = \sum_{i \in \mathcal{N}} C_i \mathbf{f}_i^{distill} \alpha_i T_i, \quad \text{where } T_i = \prod_{j=1}^{i-1} (1 - C_j \alpha_j). \label{eq:rendering}
\end{equation}
Here, $\mathbf{F}^{distill}(\mathbf{p})$ denotes the rendered distilled semantic feature at pixel $\mathbf{p}$, obtained by accumulating per-primitive features along the ray. By coupling feature accumulation with $C_i$, the field reduces semantic smearing (P1) and anchors semantic evidence primarily to stable regions.

\vspace{1mm}
\subsubsection{Stabilized View Synthesis for Spatial Reasoning}
Once optimized, $\mathcal{F}_{app}$ functions as a confidence-weighted rendering engine for downstream spatial reasoning.
Instead of feeding raw, gait-affected camera frames to the perception stack, we render stabilized views $I_{stable}$ from smoothed virtual trajectories.
These rendered views reduce gait-induced artifacts before object detection and the extracting spatial relation, providing more stable evidence for subsequent Spatial Field construction.
A later discrepancy-computation step uses the node positions, semantic features, and relation sets extracted from these views to determine whether a local memory update is needed.

\subsection{Spatial Field Construction ($\mathcal{F}_{spat}$)}
\label{sec:spatial_field}

The Spatial Field $\mathcal{F}_{spat}$ serves as the topological memory and reasoning layer of MIF.
While $\mathcal{F}_{app}$ provides confidence-weighted visual evidence, $\mathcal{F}_{spat}$ abstracts the scene into a hierarchical graph $\mathcal{M}_{spat} = (\mathcal{V}, \mathcal{E})$ for high-level task reasoning.
Its function is twofold: (1) grounding linguistic concepts into physical space using stabilized evidence, and (2) detecting likely Map-Reality Mismatches (P2) by comparing local and stored graph structure.

\subsubsection{Graph Generation via Stabilized Re-perception}
Direct object detection and semantic parsing on raw humanoid camera streams can suffer from segmentation jitter under gait-induced viewpoint changes.
We therefore employ a stabilized re-perception strategy.
We query the optimized $\mathcal{F}_{app}$ to render confidence-weighted stabilized views $\{I_{stable}\}$.
These rendered inputs reduce gait-induced artifacts before object detection and relation extraction, providing more stable evidence for scene-graph construction.
Formally, each node $v_i \in \mathcal{V}$ is represented as $v_i = \{ \mathcal{S}_i, \Omega_i, \mathcal{A}_i, \mathbf{c}_i \}$, containing a semantic caption $\mathcal{S}_i$, room affiliation $\mathcal{A}_i$, node reliability $\Omega_i$, and 3D centroid $\mathbf{c}_i$ computed via confidence-weighted back-projection:

\begin{equation}
    \mathbf{c}_i = \frac{\sum_{j \in \mathcal{N}_i} C_j \alpha_j \mathbf{x}_j}{\sum_{j \in \mathcal{N}_i} C_j \alpha_j} \label{eq:grounding} ,
\end{equation}
where $\mathbf{x}_j$ denotes the mean position of the $j$-th Gaussian primitive associated with node $v_i$.
We define the node-level reliability score as $\Omega_i = \langle C_j \rangle_{j \in \mathcal{N}_i}$, where $\langle \cdot \rangle$ denotes the arithmetic mean over associated Gaussians.
This score acts as a reliability filter: detections supported mainly by low-confidence primitives receive low $\Omega_i$, reducing the chance that gait-induced artifacts or unstable semantic proposals become persistent graph nodes.

\subsubsection{Detecting Map-Reality Mismatch via Discrepancy $\mathcal{D}$}
To address  {P2}, the system must separate transient sensing artifacts from persistent environmental changes.
We implement a closed-loop Interaction and Adaptation Mechanism that monitors a multi-modal discrepancy score $\mathcal{D}$.
The agent compares a live local subgraph $\mathcal{M}_{loc}$ against the stored global graph $\mathcal{M}_{spat}$.
For a matched local-global node pair, the node-level discrepancy $\delta_i$ is defined as:

\begin{equation}
    \delta_i = \Omega_i \cdot \left( w_{pos} \|\mathbf{c}_i^{loc} - \mathbf{c}_j^{glob}\|_2 + w_{sem} (1 - \cos(\hat{\mathbf{f}}_i, \hat{\mathbf{f}}_j)) \right). \label{eq:delta_calc}
\end{equation}
Here, $\mathbf{c}_i^{loc}$ and $\mathbf{c}_j^{glob}$ are matched local/global centroids, $\hat{\mathbf{f}}_i$ and $\hat{\mathbf{f}}_j$ are normalized semantic features, $w_{pos}$ and $w_{sem}$ weight geometric and semantic drift.
This reduces the influence of low-confidence nodes on change detection, limiting false positives caused by motion blur or gait-induced artifacts (P1).
The total discrepancy $\mathcal{D}$ aggregates node-level shifts and relational changes:

\begin{equation}
    \mathcal{D} = \frac{1}{|\mathcal{V}_{loc}|} \sum_{i \in \mathcal{V}_{loc}} \delta_i + w_{rel} \frac{|\mathcal{E}_{loc} \Delta \mathcal{E}_{spat}|}{|\mathcal{E}_{loc} \cup \mathcal{E}_{spat}|}. \label{eq:total_discrepancy}
\end{equation}
Here, $\mathcal{V}_{loc}$ averages local node discrepancies, while $\mathcal{E}_{loc}$ and $\mathcal{E}_{spat}$ are live and stored edge sets; $\Delta$ and $|\cdot|$ denote symmetric difference and cardinality, with $w_{rel}$ weighting relational inconsistency.
When $\mathcal{D} > \tau$, the system treats the local region as a likely Map-Reality Mismatch and selectively updates only the inconsistent region using accumulated multi-view evidence.
Implementation-level definitions of the three fields and the partial graph-update procedure are provided in Sec.~\ref{sec:algo_workflow} and Algorithm~\ref{alg:main_loop}.

\subsection{Geometry Field Construction ($\mathcal{F}_{geom}$)}
\label{sec:geometry_field}
The Geometry Field $\mathcal{F}_{geom}$ serves as the interaction-safety layer of MIF.
Although $\mathcal{F}_{app}$ and $\mathcal{F}_{spat}$ enable semantic grounding and topological reasoning, their sparse or implicit geometry is insufficient for selecting terminal poses for manipulation.
Safe humanoid manipulation benefits from explicit object surfaces that can support collision and reachability checks.
To bridge this gap, $\mathcal{F}_{geom}$ functions as an on-demand module that recovers object-centric meshes from stable visual evidence and uses them to evaluate Interaction Pose Safety (IPS).

\subsubsection{Task-Local View Selection for Interaction Geometry}
To reduce geometric ambiguity before interaction, the agent employs a task-local active gaze strategy around the target object.
This process samples candidate viewpoints in the local neighborhood of the object on a local hemisphere and ranks them by a geometric observation utility for the object's interaction surfaces.
We rank candidate viewpoints using the task-local utility $Q(k)$:

\begin{equation}
    Q(k) = \exp( -\| \mathbf{c}_i - \mathbf{p}_k \|_2^2 / 2\sigma_d^2 ) \cdot (\mathbf{d}_k \cdot \mathbf{v}_{ik})^\gamma \cdot \frac{1}{|\mathcal{N}_i|} \sum_{j \in \mathcal{N}_i} \Omega_j, \label{eq:qk}
\end{equation}
where $\mathbf{p}_k$ and $\mathbf{d}_k$ denote the position and optical-axis direction of candidate view $k$, and $\mathbf{v}_{ik} = (\mathbf{c}_i - \mathbf{p}_k) / \|\mathbf{c}_i - \mathbf{p}_k\|_2$ is the normalized direction from the camera center to the object centroid $\mathbf{c}_i$.
The distance term favors nearby views, while $(\mathbf{d}_k \cdot \mathbf{v}_{ik})^\gamma$ favors foveal alignment between the optical axis and the object center.
The confidence term favors views that observe high-reliability Gaussian neighborhoods, so the generative model is conditioned on stable evidence rather than gait-corrupted observations.
This links active perception to interaction feasibility by improving the observability of surfaces that are relevant to the intended manipulation.

\subsubsection{Conditional Flow Matching for 3D Mesh Recovery}
To infer partially observed object surfaces (\emph{e.g.}, the occluded back side), we employ a Flow Matching-based~\cite{lipman2022flow} reconstruction model~\cite{chen2025sam}.
We formulate geometric recovery as a conditional generation problem.
Specifically, the model learns a time-dependent vector field that transports a Gaussian noise distribution toward the target mesh distribution conditioned on the selected observation.

This generation is conditioned on the selected view ranked by the task-local utility $Q(k)$.
By conditioning on this high-confidence visual evidence, the model recovers an object-centric mesh that is consistent with the denoised observations from $\mathcal{F}_{app}$ and can be used by the IPS check.

\subsubsection{Asynchronous Grounding and IPS Verification}
To avoid blocking navigation during generative inference, we implement an asynchronous grounding pipeline.
While the robot approaches the target, the selected-view mesh is synthesized asynchronously and aligned to the Gaussian support using scale-invariant ICP:

\begin{equation}
    \min_{\mathbf{R}, \mathbf{t}, s} \sum_{m=1}^M \rho \left( \|s \mathbf{R} \mathbf{x}_{gen}^m + \mathbf{t} - \mathbf{x}_{app}^{NN(m)}\| \right) .
\end{equation}
Here, ICP estimates the rotation $\mathbf{R}$, translation $\mathbf{t}$, and scale $s$ that align each generated mesh point $\mathbf{x}_{gen}^m$ to its nearest Gaussian-support point $\mathbf{x}_{app}^{NN(m)}$, with robust loss $\rho$ reducing outlier influence.
The registered mesh $\mathcal{M}_{gen}$ provides dense local geometry for evaluating the final standing pose $\mathbf{p}_{se3}$.

\vspace{1mm}
\subsubsection{Interaction Pose Safety}
The IPS score $S_{IPS}(\mathbf{p}_{se3})$ is an operational safety criterion for checking whether the terminal pose is suitable for manipulation:

\begin{equation}
    S_{IPS} = \mathbf{I}_{col}(\mathbf{p}_{se3}, \mathcal{M}_{gen}) \land \mathbf{I}_{ik}(\mathbf{p}_{se3}, \mathbf{t}_{obj}) \land \mathbf{I}_{stab}(\mathbf{p}_{se3}) ,
\end{equation}
where:
\begin{itemize}
    \item $\mathbf{I}_{col}$ ( {Geometric Safety}): evaluates the signed distance to $\mathcal{M}_{gen}$ and checks whether the collision margin exceeds $\delta_{safe}$.
    \item $\mathbf{I}_{ik}$ ( {Kinematic Reachability}): returns 1 if a valid collision-free inverse-kinematics solution exists for reaching the target point $\mathbf{t}_{obj}$.
    \item $\mathbf{I}_{stab}$ ( {Postural Stability}): returns 1 if the projected center of mass (CoM) lies within the support polygon $\mathcal{P}_{sup}$, defined by the foot contact region at pose $\mathbf{p}_{se3}$.
\end{itemize}
Thus, navigation is considered complete only when the robot reaches a pose that is collision-free, kinematically reachable, and stable for the subsequent manipulation task, rather than merely arriving near the target coordinate.

\subsection{Language-Conditioned Navigation and Adaptation}
\label{sec:navigation}
The final stage converts language-conditioned goals into navigation and verification actions.
Using the hierarchical structure of MIF, the system selects semantically grounded targets, verifies geometry-aware terminal poses, and updates local memory when persistent scene changes are detected.

\subsubsection{Hierarchical Semantic Grounding}
Following a hierarchical search strategy similar to~\cite{Werby-RSS-24}, the agent parses a natural-language instruction (\emph{e.g.}, \textit{``Grasp the blue mug on the table''}) into a $\langle \text{Region}, \text{Landmark}, \text{Object} \rangle$ triplet using a VLM~\cite{hurst2024gpt}.
The system sequentially queries $\mathcal{F}_{spat}$ to constrain the search space by localizing the room, identifying the landmark, and retrieving the target object node $v_i$.
The object's centroid $\mathbf{c}_i$, computed by Eq.~\ref{eq:grounding}, provides the initial navigation goal, while the final stance is verified by IPS before interaction.

\subsubsection{Stability-Aware Execution and Safety Verification}
\begin{figure}[t]
    \centering
    \includegraphics[width=0.90\columnwidth]{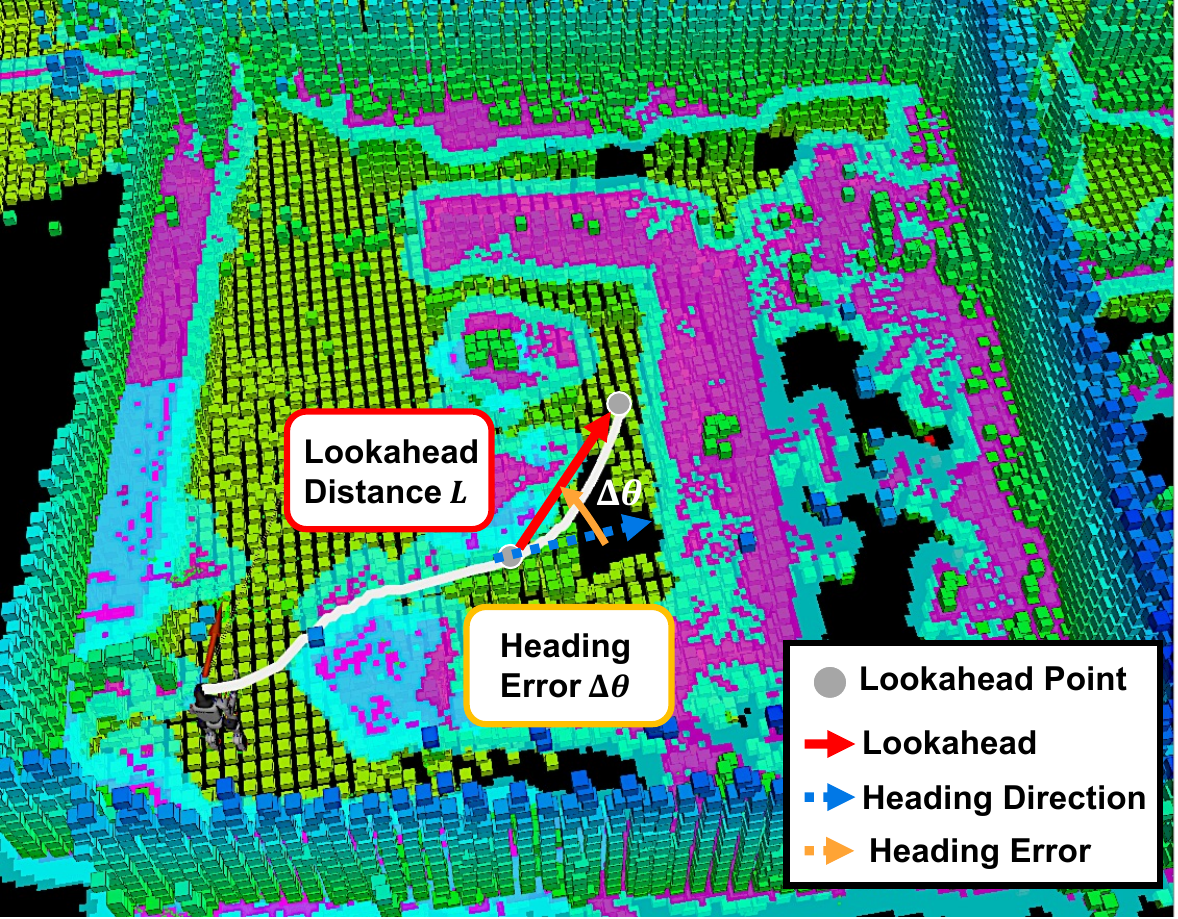}
    \caption{\textbf{Humanoid navigation control via Pure Pursuit.} The schematic illustrates the geometric tracking logic where the Unitree-G1 robot targets a lookahead point at distance $L$ along the planned trajectory. The heading error $\Delta\theta$ is used to compute the required curvature $\kappa = 2\sin(\Delta\theta)/L$. The overlaid velocity curve demonstrates the adaptive scaling mechanism, where $L$ is dynamically adjusted to suppress oscillations and maintain humanoid balance during high-curvature maneuvers.}
    \label{fig:navigation_control}
\end{figure}

To manage humanoid locomotion constraints, we use a stability-aware local planner that prioritizes center-of-mass (CoM) stability when tracking the planned path.
The planner scales velocity according to angular tracking error to reduce gait-induced oscillations and preserve perception quality during motion, as shown in Fig.~\ref{fig:navigation_control}.

Navigation completion is defined by the IPS criterion rather than simple coordinate arrival.
Before manipulation, the agent evaluates the candidate stance against the registered geometry $\mathcal{M}_{gen}$.
If the IPS checks indicate collision risk or insufficient reachability, the local planner adjusts the stance within the nearby feasible region and re-evaluates $S_{IPS}$.

\subsubsection{Closed-Loop Memory Evolution}
During navigation, the Interaction and Adaptation Mechanism uses new observations to validate the local consistency of the stored memory.
The discrepancy score $\mathcal{D}$ distinguishes between two operational cases:

\begin{itemize}
    \item  {Transient Discrepancy:} temporary obstacles (\emph{e.g.}, moving people) trigger local avoidance without updating the persistent scene memory.
    \item  {Structural Mismatch:} a persistent increase in $\mathcal{D}$ above $\tau$ indicates a likely map-reality conflict (\emph{e.g.}, relocated furniture). The system then scans the discrepant region, updates only the locally inconsistent parts of $\mathcal{F}_{app}$ and $\mathcal{F}_{spat}$, and re-grounds the query in the updated topology.
\end{itemize}

\section{EXPERIMENTAL EVALUATION}
\label{sec:experiments}
The experiments evaluate whether MIF enables a humanoid robot to maintain, query, and revise a multi-modal scene memory during real-world navigation and interaction.
We organize the evaluation around the two vulnerabilities introduced earlier: locomotion-induced semantic-geometric distortion (P1) and map-reality mismatch under scene changes (P2).
The main paper reports the following three questions, with additional implementation details and visualizations provided in the supplementary material.

\begin{itemize}
    \item \textbf{Q1 (Solving P1):} Can the Appearance and Spatial Fields preserve semantic grounding under bipedal camera jitter and fast walking?
    \item \textbf{Q2 (Enabling Interaction):} Does target-centered geometry reconstruction improve IPS-verified interaction feasibility compared with point-cloud and static-fusion alternatives?
    \item \textbf{Q3 (Solving P2):} Can the discrepancy-triggered adaptation loop distinguish persistent scene changes from transient sensing artifacts and update only inconsistent local memory?
\end{itemize}

Because the system spans mapping, semantic grounding, interaction safety, and memory adaptation, we report the task-level evidence in the main paper and keep extended qualitative sequences and parameter settings in the supplementary.

\subsection{Experimental Setup}
We conduct real-world experiments in an approximately $100~\text{m}^2$ office environment containing more than 100 object categories, where objects can be relocated, removed, or newly introduced after mapping.
The full ROS1~\cite{quigley2009ros} system runs on a centralized workstation with an NVIDIA RTX 4090 and communicates with a Unitree G1 humanoid during navigation and interaction trials.
All compared methods use the same SLAM backend where applicable, so the reported differences mainly reflect memory construction, semantic grounding, geometry reconstruction, and adaptation.
The supplementary material provides additional system implementation details, hyperparameter settings, and qualitative visualization results.

The baselines are selected according to each question rather than treated as a single universal set.
\begin{itemize}
    \item For semantic-feature baselines, we include Feature-Splatting~\cite{qiu2024featuresplatting}, LangSplat~\cite{qin2024langsplat}, LatentBKI~\cite{wilson2025latentbki}, and MIF variants with different feature dimensions.
    \item For scene-graph and dynamic-memory baselines, we compare with HOV-SG~\cite{Werby-RSS-24}, ConceptGraphs~\cite{gu2024conceptgraphs} with periodic rescanning, and Khronos~\cite{schmid2024khronos} when evaluating adaptation after scene changes, without allowing a separate global rescan before task execution.
    \item For rendering robustness under humanoid walking, we further compare against Deblur-GS and PUP-3DGS using slow- and fast-walk image-quality metrics.
\end{itemize}

\subsection{Robust Semantic Grounding under Distortion (Q1)}
\label{sec:exp_semantic_grounding}
This experiment evaluates whether confidence-gated rendering in $\mathcal{F}_{app}$ provides stable evidence for semantic localization and graph construction in $\mathcal{F}_{spat}$ under P1.
The Unitree G1 receives open-vocabulary object queries, such as ``Find the water bottle'', and must localize the target in the current scene memory.

\subsubsection{Quantitative Results}
Following HOV-SG, a query is counted as successful when the predicted object matches the ground-truth category and its centroid is within $1.0\,\text{m}$ of the target.
Because interaction requires tighter spatial accuracy than goal-level navigation, we also report mean distance error (MDE).
Table~\ref{tab:semantic_results} shows that MIF improves task-level semantic localization under humanoid motion, with higher success rate and lower MDE than HOV-SG and Feature-Splatting in this setup.
The lower MDE suggests that confidence-gated integration suppresses locomotion-induced smearing that otherwise shifts semantic anchors away from object centers.

\begin{table}[h]
\centering
\footnotesize
\caption{Quantitative results of semantic grounding under locomotion-induced distortion.}
\label{tab:semantic_results}
\begin{tabular*}{\linewidth}{@{\extracolsep{\fill}}lccc@{}}
\toprule
\textbf{Method} & \textbf{Success Rate} & \textbf{MDE (m)} & \textbf{Latency (s)} \\ \midrule
Feature-Splatting~\cite{qiu2024featuresplatting} & 65\% & 0.42 & 3.2 \\
HOV-SG~\cite{Werby-RSS-24} & 74\% & 0.35 & 2.8 \\
\textbf{MIF (Ours)} & \textbf{92\%} & \textbf{0.18} & \textbf{1.6} \\ \bottomrule
\end{tabular*}
\end{table}
To isolate rendering robustness, Table~\ref{tab:conf} compares MIF with Deblur-GS and PUP-3DGS under slow and fast walking.
\begin{table}[h]
\centering
\footnotesize
\caption{Rendering results under slow (S-) and fast (F-) walk.}
\label{tab:conf}
\begin{tabular*}{\linewidth}{@{\extracolsep{\fill}}lcccc@{}}
\toprule
\textbf{Method} & \textbf{S-PSNR$\uparrow$} & \textbf{S-SSIM$\uparrow$} & \textbf{F-PSNR$\uparrow$} & \textbf{F-SSIM$\uparrow$} \\ \midrule
w/o Confidence & 28.4 & 0.88 & 21.6 & 0.72 \\
Deblur-GS~\cite{oh2024deblurgs} & 25.14 & 0.80 & 24.67 & 0.70 \\
PUP-3DGS~\cite{hanson2025pup} & 20.44 & 0.67 & 17.89 & 0.53 \\
\textbf{MIF $C_i$ (Ours)} & \textbf{31.2} & \textbf{0.93} & \textbf{29.8} & \textbf{0.89} \\ \bottomrule
\end{tabular*}
\end{table}

MIF is not using confidence only for visual quality; the same reliability estimate is used to gate map integration and reduce unstable evidence before graph construction.

Table~\ref{tab:render} further shows that scene graphs constructed from stabilized rendered views achieve better object localization than those constructed from raw walking frames, especially under fast walking.
\begin{table}[h]
\centering
\footnotesize
\caption{Semantic object localization on scene graphs constructed from rendered vs. raw frames.}
\label{tab:render}
\begin{tabular*}{\linewidth}{@{\extracolsep{\fill}}lccc@{}}
\toprule
\textbf{Input Source} & \textbf{Recall$\uparrow$} & \textbf{Precision$\uparrow$} & \textbf{mAP$\uparrow$} \\ \midrule
Raw frames (Slow) & 0.87 & 0.91 & 0.84 \\
Raw frames (Fast) & 0.61 & 0.79 & 0.58 \\
\textbf{MIF Rendered (Slow)} & \textbf{0.94} & \textbf{0.95} & \textbf{0.93} \\
\textbf{MIF Rendered (Fast)} & \textbf{0.92} & \textbf{0.94} & \textbf{0.91} \\
Static camera (Ref.) & $\sim$1.00 & $\sim$1.00 & $\sim$1.00 \\ \bottomrule
\end{tabular*}
\end{table}

\subsection{High-fidelity Geometry for Interaction Safety (Q2)}
\label{sec:exp_navigation}

This experiment evaluates the transition from object semantic grounding and navigation to IPS-verified interaction.
We test whether the Geometry Field provides a safer physical proxy for terminal interaction poses than sparse point-cloud or static-fusion alternatives.

\subsubsection{The Necessity of High-fidelity Geometry}
Compared with raw point-cloud planning, MIF reduces IPS failure from 38\% to 6\% in our trials, mainly because continuous surfaces expose feasible clearances more reliably than sparse point clouds formed by Gaussian centers.
The point-cloud baseline often misses surface connectivity, while the static mesh baseline can preserve locomotion-induced artifacts that occlude otherwise feasible workspaces.

For safety, sparse point clouds can make IPS checks unreliable: without continuous surface connectivity, they provide incomplete normal and boundary information for manipulation, causing $\mathbf{I}_{col}$ to mark some risky poses as feasible.
The generated watertight meshes reduce this failure mode and make the collision term in IPS better aligned with physical contact constraints.

\subsubsection{Interaction Performance}
We count an interaction trial as successful only when the robot arrives and the terminal pose satisfies $S_{IPS}=1$.
MIF achieves a mean terminal position error of $0.12\,\text{m}$ in these trials.
This precision supports the reachability check $\mathbf{I}_{ik}$, since larger terminal errors can move the target outside the feasible manipulation workspace..
The active viewpoint score $Q(k)$ is reported as a task-local geometric observation utility: it selects views with better distance, alignment, and high-confidence Gaussian support before mesh reconstruction.
These cleaner conditioning views support Flow-Matching reconstruction and improve IPS performance as part of the overall system.

\vspace{1mm}
\subsubsection{Baseline Comparison Analysis}
To examine these gains, we compared MIF against two ablative baselines:
\begin{itemize}
    \item Pt-Cloud Baseline: Pt-Cloud plans on sparse Gaussian centroids; without surface connectivity, collision checks can be unreliable around gaps.
    \item Static Mesh Baseline: Static Mesh uses TSDF fusion without the confidence gate; locomotion artifacts can be integrated into the mesh and reduce feasible terminal-pose estimates.
\end{itemize}
As shown in Table~\ref{tab:nav_results}, MIF attains 94\% IPS success with no observed collisions, suggesting that confidence-gated mapping and on-demand geometry are complementary for interaction safety.

\begin{table}[h]
\centering
\footnotesize
\caption{Interaction Feasibility Comparison. We compare our generative approach against a Pt-Cloud Baseline and a Static Mesh Baseline.}
\label{tab:nav_results}
\begin{tabular*}{\linewidth}{@{\extracolsep{\fill}}lccc@{}}
\toprule
\textbf{Method} & \textbf{IPS Success (\%)} & \textbf{Nav. Err. (m)} & \textbf{Collision Rate} \\ \midrule
Pt-Cloud & 62\% & 0.28 & 32\% \\
Static Mesh & 78\% & 0.22 & 14\% \\
\textbf{MIF (Ours)} & \textbf{94\%} & \textbf{0.12} & \textbf{0\%} \\ \bottomrule
\end{tabular*}
\end{table}

\subsection{Dynamic Adaptation to Map-Reality Mismatches (Q3)}
\label{sec:exp_adaptation}

\begin{figure*}[t]
    \centering
    \includegraphics[width=0.9\textwidth]{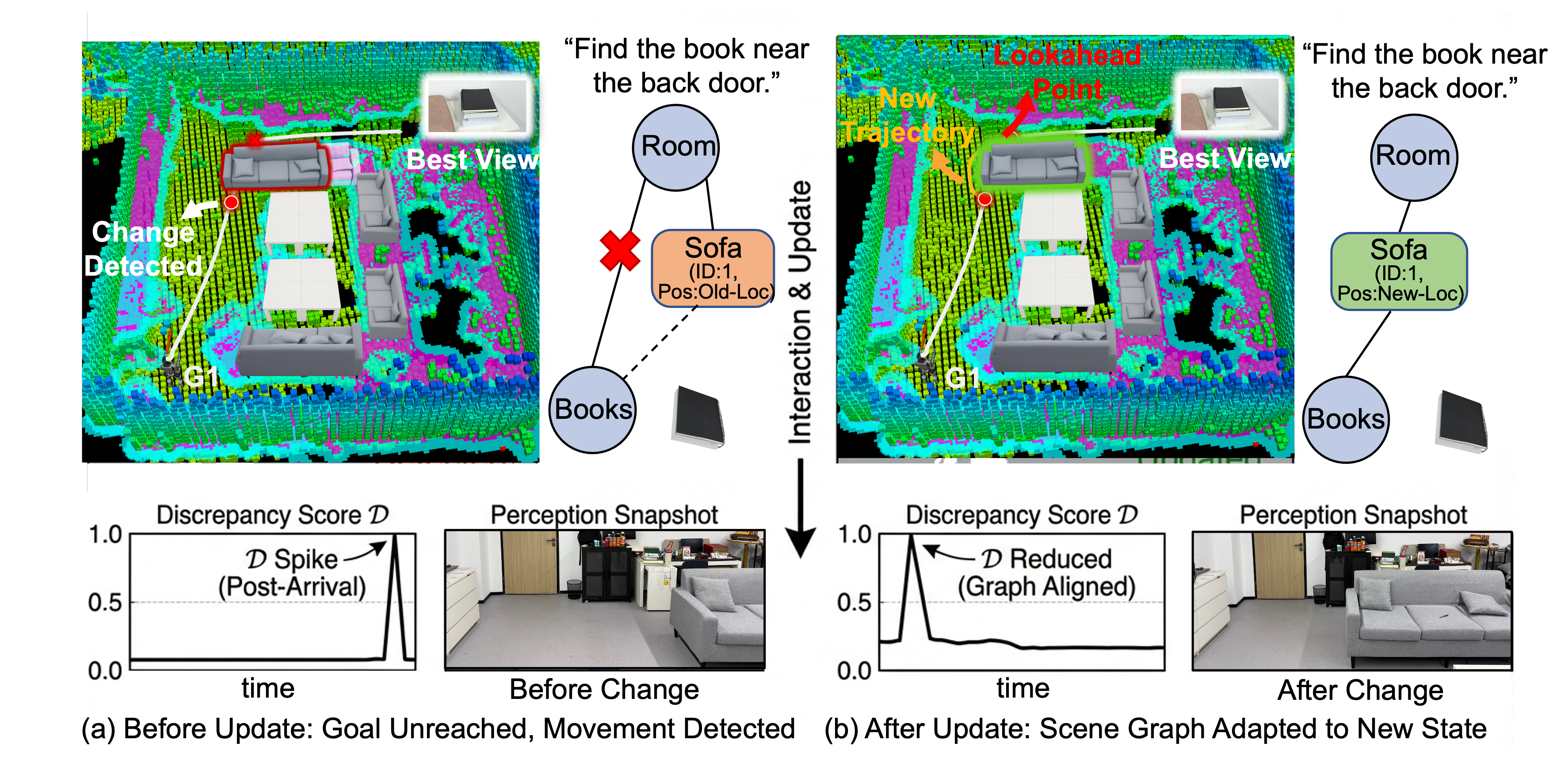}
    \caption{\textbf{Dynamic adaptation of the hierarchical scene graph to environmental changes.} (a) \textbf{Detection}: Upon arriving at the goal, the G1 robot detects a structural discrepancy (\emph{e.g.}, a moved sofa) that contradicts its prior memory, triggering an increase in the discrepancy score $\mathcal{D}$. (b) \textbf{Resolution}: MIF initiates a local update of the Spatial Field with fresh observations. The updated graph better aligns with the physical reality, reducing $\mathcal{D}$ and improving consistency for subsequent tasks.}
    \label{fig:adaptation}
\end{figure*}

This experiment evaluates whether the discrepancy-triggered adaptation loop mitigates P2 after the initial map becomes stale. 
After initial mapping, we introduce three types of scene changes: object relocation, removal, and addition.
For relocation, the target is moved by more than $1.5\,\text{m}$, and the robot must adapt during task execution without a separate global rescan after the scene change.

\subsubsection{Evaluation Protocol}
A trial is successful only if the robot either reaches the target's current physical location with an IPS-valid final pose or correctly reports that the target has been removed.
Arriving at obsolete memory coordinates is counted as failure.
All methods start from the same initial map and post-change task condition, and no method is allowed to perform a separate global rescan before task execution. During execution, each method uses the observations produced by its own navigation or adaptation strategy.

\subsubsection{Results Analysis}
During navigation, MIF compares the live local graph with the stored Spatial Field using $\mathcal{D}$, which aggregates positional, semantic, and relational evidence. Fig.~\ref{fig:adaptation} illustrates how a detected discrepancy triggers a local Spatial Field update after scene changes.
Table~\ref{tab:tau} reports the ROC-based threshold selection used to set $\tau=0.45$, showing the trade-off between change recall and false-positive updates.

\begin{table}[h]
\centering
\caption{ROC-based threshold selection in real scenarios.}
\label{tab:tau}
\footnotesize
\begin{tabular*}{\linewidth}{@{\extracolsep{\fill}}cccc@{}}
\toprule
\textbf{Threshold $\bm{\tau}$} & \textbf{True Positive Rate$\uparrow$} & \textbf{False Positive Rate$\downarrow$} & \textbf{F1$\uparrow$} \\ \midrule
0.25 & 99.2\% & 18.7\% & 0.817 \\
0.35 & 97.8\% & 9.3\% & 0.893 \\
\textbf{0.45 (Ours)} & \textbf{95.1\%} & \textbf{4.2\%} & \textbf{0.934} \\
0.55 & 89.6\% & 1.8\% & 0.913 \\
0.65 & 78.3\% & 0.7\% & 0.866 \\ \bottomrule
\end{tabular*}
\end{table}

This threshold is fixed for all scene-change trials. Table~\ref{tab:adaptation_comparison} then compares static-memory, periodic-rescan, dynamic scene-graph, and MIF variants under the same setting. HOV-SG has low success when the target is no longer at the stored coordinates, while ConceptGraphs+Rescan and Khronos provide stronger baselines for evaluating whether scene memories can be revised after changes.
For MIF, we report two stages to separate performance without memory revision from the final performance after memory revision is enabled. MIF (Initial) disables the local memory update mechanism and only uses local observations near the obsolete memory location; it can still succeed when the relocated object remains visible or close enough for local perception and planning. MIF (Post-Update) reports the final success rate of the full system, including the initially solved cases and the additional cases recovered after $\mathcal{D}>\tau$ triggers active scanning, local graph revision, and replanning.

\begin{table}[h]
\centering
\footnotesize
\caption{Scene-change task success without a separate global rescan before task execution. Reloc/Remov/Addit denote relocated ($>1.5$m), removed, and newly added objects. MIF (Initial) uses local observations near the obsolete memory location without local memory revision, while MIF (Post-Update) reports final task success after active scanning, local graph revision, and replanning are enabled.}
\label{tab:adaptation_comparison}
\begin{tabular*}{\linewidth}{@{\extracolsep{\fill}}lccc@{}}
\toprule
\textbf{Method} & \textbf{Reloc SR$\uparrow$} & \textbf{Remov SR$\uparrow$} & \textbf{Addit SR$\uparrow$} \\ \midrule
HOV-SG~\cite{Werby-RSS-24} & 12\% & 10\% & 0\% \\
ConceptGraphs + Rescan~\cite{gu2024conceptgraphs} & 38\% & 35\% & 12\% \\
Khronos~\cite{schmid2024khronos} & $\sim$18\% & 61\% & $\sim$12\% \\
MIF (Initial) & 64\% & 68\% & 42\% \\
\textbf{MIF (Post-Update)} & \textbf{94\%} & \textbf{98\%} & \textbf{86\%} \\ \bottomrule
\end{tabular*}
{\raggedright $^*$The two-session protocol first builds the initial map, then changes the scene and evaluates task completion during post-change navigation. The robot is not required to exactly repeat the initial mapping trajectory.\par}
\vspace{-3mm}
\end{table}

\subsection{System Efficiency}

\label{sec:exp_performance}
We finally report computational cost because online humanoid deployment requires a balanced allocation of computational resources, with sufficient headroom for perception, language grounding, geometry generation, and control.

Table~\ref{tab:semantic_embedding} compares MIF's 32D and 64D variants with LangSplat, LatentBKI, and FeatSplat under slow and fast walking.

\begin{table}[h]
\centering
\footnotesize
\caption{Semantic localization under slow (S-) and fast (F-) walk conditions, memory usage, and latency.}
\label{tab:semantic_embedding}
\begin{tabular*}{\linewidth}{@{\extracolsep{\fill}}lcccc@{}}
\toprule
\textbf{Method} & \textbf{VRAM$\downarrow$\,(GB)} & \textbf{Latency$\downarrow$\,(s)} & \textbf{S-SR$\uparrow$} & \textbf{F-SR$\uparrow$} \\ \midrule
LangSplat~\cite{qin2024langsplat} & 18.45 & 2.3 & 95\% & 72\% \\
LatentBKI-64D~\cite{wilson2025latentbki} & 23.5 & 7.1 & \textbf{98\%} & 74\% \\
FeatSplat-32D~\cite{qiu2024featuresplatting} & 4.8 & 3.2 & 81\% & 59\% \\
\textbf{MIF 64D (Ours)} & 18.0 & 4.6 & 92\% & \textbf{88\%} \\
\textbf{MIF 32D (Ours)} & \textbf{$<$4.0} & \textbf{1.6} & 90\% & 86\% \\ \bottomrule
\end{tabular*}
\end{table}
    
In practice, $\mathcal{F}_{app}$ supports real-time incremental updates at 22 FPS, while the asynchronous generation of $\mathcal{F}_{geom}$ takes around 6.2s and is scheduled during approach or verification rather than blocking low-level control. The 32D semantic feature field further reduces VRAM usage to below 4GB for the tested office map, leaving capacity for the VLM/LLM and mesh-generation modules on the RTX 4090.

\section{CONCLUSION}
\label{sec:conclusion}
We presented Multi-modal Interactive Field (MIF), a humanoid-oriented system for manipulation-safety-aware navigation in changing indoor environments. MIF integrates a confidence-aware Appearance Field, a topology-aware Spatial Field, and an object-centric Geometry Field to mitigate locomotion-induced semantic-geometric distortion (P1) and map-reality mismatch under scene changes (P2). On a Unitree G1 humanoid, MIF achieves 92\% semantic-grounding success, 94\% IPS success with no observed collisions, and final adaptation success rates of 94\%, 98\%, and 86\% for relocation, removal, and addition. These results suggest that coupling reliability estimation, discrepancy-based local memory revision, and target-centered geometry verification improves humanoid interaction in non-static environments. Remaining limitations include several-second geometry generation latency and reliance on a centralized RTX 4090 workstation; future work will reduce latency, move more computation onboard, and develop multiplex interactive learning so that visual, spatial, geometric, and human feedback can jointly refine the robot's memory and adaptation policy over repeated interactions.

\section*{Acknowledgments}
This work is jointly supported by the National Natural Science Foundation of China (No.62373009), and Guangdong S\&T Program (No.2024B0101050002). We specifically thank Jin Jin for his critical insights into the overall methodology and extensive revisions of the manuscript.

\bibliographystyle{plainnat}
\bibliography{references}

@INPROCEEDINGS{Werby-RSS-24, 
  AUTHOR    = {Abdelrhman Werby AND Chenguang Huang AND Martin Büchner AND Abhinav Valada AND Wolfram Burgard}, 
  TITLE     = {Hierarchical Open-Vocabulary 3D Scene Graphs for Language-Grounded Robot Navigation}, 
  BOOKTITLE = {Robotics: Science and Systems (RSS)}, 
  YEAR      = {2024}
}

@inproceedings{engelmann2024opennerf,
  title={Opennerf: Open set 3d neural scene segmentation with pixel-wise features and rendered novel views},
  author={Engelmann, Francis and Manhardt, Fabian and Niemeyer, Michael and Tateno, Keisuke and Tombari, Federico},
  booktitle={International Conference on Learning Representations (ICLR)},
  volume={2024},
  pages={8396--8407},
  year={2024}
}

@article{kobayashi2022decomposing,
  title={Decomposing nerf for editing via feature field distillation},
  author={Kobayashi, Sosuke and Matsumoto, Eiichi and Sitzmann, Vincent},
  journal={Advances in Neural Information Processing Systems (NeurIPS)},
  volume={35},
  pages={23311--23330},
  year={2022}
}

@inproceedings{shi2024language,
  title={Language embedded 3d gaussians for open-vocabulary scene understanding},
  author={Shi, Jin-Chuan and Wang, Miao and Duan, Hao-Bin and Guan, Shao-Hua},
  booktitle={Proceedings of the IEEE/CVF Conference on Computer Vision and Pattern Recognition (CVPR)},
  pages={5333--5343},
  year={2024}
}

@inproceedings{peng2023openscene,
  title={Openscene: 3d scene understanding with open vocabularies},
  author={Peng, Songyou and Genova, Kyle and Jiang, Chiyu and Tagliasacchi, Andrea and Pollefeys, Marc and Funkhouser, Thomas and others},
  booktitle={Proceedings of the IEEE/CVF Conference on Computer Vision and Pattern Recognition (CVPR)},
  pages={815--824},
  year={2023}
}

@inproceedings{kirillov2023segment,
  title={Segment anything},
  author={Kirillov, Alexander and Mintun, Eric and Ravi, Nikhila and Mao, Hanzi and Rolland, Chloe and Gustafson, Laura and Xiao, Tete and Whitehead, Spencer and Berg, Alexander C and Lo, Wan-Yen and others},
  booktitle={Proceedings of the IEEE/CVF International Conference on Computer Vision (ICCV)},
  pages={4015--4026},
  year={2023}
}

@article{hurst2024gpt,
  title={Gpt-4o system card},
  author={Hurst, Aaron and Lerer, Adam and Goucher, Adam P and Perelman, Adam and Ramesh, Aditya and Clark, Aidan and Ostrow, AJ and Welihinda, Akila and Hayes, Alan and Radford, Alec and others},
  journal={arXiv preprint arXiv:2410.21276},
  year={2024}
}

@inproceedings{lipman2022flow,
    title={Flow Matching for Generative Modeling},
    author={Yaron Lipman and Ricky T. Q. Chen and Heli Ben-Hamu and Maximilian Nickel and Matthew Le},
    booktitle={The Eleventh International Conference on Learning Representations (ICLR)},
    year={2023}
}

@article{kerbl20233d,
  title={3D Gaussian splatting for real-time radiance field rendering},
  author={Kerbl, Bernhard and Kopanas, Georgios and Leimk{\"u}hler, Thomas and Drettakis, George},
  journal={ACM Trans. Graph.},
  volume={42},
  number={4},
  pages={139--1},
  year={2023}
}

@article{chaplot2020object,
  title={Object goal navigation using goal-oriented semantic exploration},
  author={Chaplot, Devendra Singh and Gandhi, Dhiraj Prakashchand and Gupta, Abhinav and Salakhutdinov, Russ R},
  journal={Advances in Neural Information Processing Systems (NeurIPS)},
  volume={33},
  pages={4247--4258},
  year={2020}
}

@inproceedings{chaplot2020learning,
  title={Learning To Explore Using Active Neural SLAM},
  author={Chaplot, Devendra Singh and Gandhi, Dhiraj and Gupta, Saurabh and Gupta, Abhinav and Salakhutdinov, Ruslan},
  booktitle={International Conference on Learning Representations (ICLR)},
  year={2020}
}

@article{chen2023not,
  title={How to not train your dragon: Training-free embodied object goal navigation with semantic frontiers},
  author={Chen, Junting and Li, Guohao and Kumar, Suryansh and Ghanem, Bernard and Yu, Fisher},
  booktitle={Robotics: Science and Systems (RSS)},
  year={2024}
}

@inproceedings{zhu2025move,
  title={Move to understand a 3d scene: Bridging visual grounding and exploration for efficient and versatile embodied navigation},
  author={Zhu, Ziyu and Wang, Xilin and Li, Yixuan and Zhang, Zhuofan and Ma, Xiaojian and Chen, Yixin and Jia, Baoxiong and Liang, Wei and Yu, Qian and Deng, Zhidong and others},
  booktitle={Proceedings of the IEEE/CVF International Conference on Computer Vision (ICCV)},
  pages={8120--8132},
  year={2025}
}

@article{yin2024sg,
  title={Sg-nav: Online 3d scene graph prompting for llm-based zero-shot object navigation},
  author={Yin, Hang and Xu, Xiuwei and Wu, Zhenyu and Zhou, Jie and Lu, Jiwen},
  journal={Advances in neural information processing systems (NeurIPS)},
  volume={37},
  pages={5285--5307},
  year={2024}
}

@inproceedings{yokoyama2024vlfm,
  title={Vlfm: Vision-language frontier maps for zero-shot semantic navigation},
  author={Yokoyama, Naoki and Ha, Sehoon and Batra, Dhruv and Wang, Jiuguang and Bucher, Bernadette},
  booktitle={IEEE International Conference on Robotics and Automation (ICRA)},
  pages={42--48},
  year={2024}
}

@inproceedings{kim2013perception,
  title={Perception-driven navigation: Active visual SLAM for robotic area coverage},
  author={Kim, Ayoung and Eustice, Ryan M},
  booktitle={IEEE International Conference on Robotics and Automation (ICRA)},
  pages={3196--3203},
  year={2013}
}

@inproceedings{xu2021invariant,
  title={Invariant EKF based 2D active SLAM with exploration task},
  author={Xu, Mengya and Song, Yang and Chen, Yongbo and Huang, Shoudong and Hao, Qi},
  booktitle={IEEE International Conference on Robotics and Automation (ICRA)},
  pages={5350--5356},
  year={2021}
}

@article{zhao2022perception,
  title={Perception-aware planning for active SLAM in dynamic environments},
  author={Zhao, Yao and Xiong, Zhi and Zhou, Shuailin and Wang, Jingqi and Zhang, Ling and Campoy, Pascual},
  journal={Remote Sensing},
  volume={14},
  number={11},
  pages={2584},
  year={2022},
  publisher={MDPI}
}

@inproceedings{gadre2023cows,
  title={Cows on pasture: Baselines and benchmarks for language-driven zero-shot object navigation},
  author={Gadre, Samir Yitzhak and Wortsman, Mitchell and Ilharco, Gabriel and Schmidt, Ludwig and Song, Shuran},
  booktitle={Proceedings of the IEEE/CVF Conference on Computer Vision and Pattern Recognition (CVPR)},
  pages={23171--23181},
  year={2023}
}

@inproceedings{nash2020polygen,
  title={Polygen: An autoregressive generative model of 3d meshes},
  author={Nash, Charlie and Ganin, Yaroslav and Eslami, SM Ali and Battaglia, Peter},
  booktitle={International conference on machine learning (ICML)},
  pages={7220--7229},
  year={2020},
  organization={PMLR}
}

@inproceedings{groueix2018papier,
  title={A papier-m{\^a}ch{\'e} approach to learning 3d surface generation},
  author={Groueix, Thibault and Fisher, Matthew and Kim, Vladimir G and Russell, Bryan C and Aubry, Mathieu},
  booktitle={Proceedings of the IEEE Conference on Computer Vision and Pattern Recognition (CVPR)},
  pages={216--224},
  year={2018}
}

@inproceedings{dai2019scan2mesh,
  title={Scan2mesh: From unstructured range scans to 3d meshes},
  author={Dai, Angela and Nie{\ss}ner, Matthias},
  booktitle={Proceedings of the IEEE/CVF Conference on Computer Vision and Pattern Recognition (CVPR)},
  pages={5574--5583},
  year={2019}
}

@inproceedings{chen2020bsp,
  title={Bsp-net: Generating compact meshes via binary space partitioning},
  author={Chen, Zhiqin and Tagliasacchi, Andrea and Zhang, Hao},
  booktitle={Proceedings of the IEEE/CVF Conference on Computer Vision and Pattern Recognition (CVPR)},
  pages={45--54},
  year={2020}
}

@inproceedings{ni2024grid,
  title={Grid: Scene-graph-based instruction-driven robotic task planning},
  author={Ni, Zhe and Deng, Xiaoxin and Tai, Cong and Zhu, Xinyue and Xie, Qinghongbing and Huang, Weihang and Wu, Xiang and Zeng, Long},
  booktitle={IEEE/RSJ International Conference on Intelligent Robots and Systems (IROS)},
  pages={13765--13772},
  year={2024}
}

@article{chalvatzaki2023learning,
  title={Learning to reason over scene graphs: a case study of finetuning gpt-2 into a robot language model for grounded task planning},
  author={Chalvatzaki, Georgia and Younes, Ali and Nandha, Daljeet and Le, An Thai and Ribeiro, Leonardo FR and Gurevych, Iryna},
  journal={Frontiers in Robotics and AI},
  volume={10},
  pages={1221739},
  year={2023},
  publisher={Frontiers Media SA}
}

@article{bavle2022situational,
  title={Situational graphs for robot navigation in structured indoor environments},
  author={Bavle, Hriday and Sanchez-Lopez, Jose Luis and Shaheer, Muhammad and Civera, Javier and Voos, Holger},
  journal={IEEE Robotics and Automation Letters},
  volume={7},
  number={4},
  pages={9107--9114},
  year={2022},
  publisher={IEEE}
}

@inproceedings{wu2021scenegraphfusion,
  title={Scenegraphfusion: Incremental 3d scene graph prediction from rgb-d sequences},
  author={Wu, Shun-Cheng and Wald, Johanna and Tateno, Keisuke and Navab, Nassir and Tombari, Federico},
  booktitle={Proceedings of the IEEE/CVF Conference on Computer Vision and Pattern Recognition (CVPR)},
  pages={7515--7525},
  year={2021}
}

@inproceedings{wald2020learning,
  title={Learning 3d semantic scene graphs from 3d indoor reconstructions},
  author={Wald, Johanna and Dhamo, Helisa and Navab, Nassir and Tombari, Federico},
  booktitle={Proceedings of the IEEE/CVF Conference on Computer Vision and Pattern Recognition (CVPR)},
  pages={3961--3970},
  year={2020}
}

@inproceedings{armeni20193d,
  title={3d scene graph: A structure for unified semantics, 3d space, and camera},
  author={Armeni, Iro and He, Zhi-Yang and Gwak, JunYoung and Zamir, Amir R and Fischer, Martin and Malik, Jitendra and Savarese, Silvio},
  booktitle={Proceedings of the IEEE/CVF International Conference on Computer Vision (ICCV)},
  pages={5664--5673},
  year={2019}
}

@inproceedings{kajita20013d,
  title={The 3D linear inverted pendulum mode: A simple modeling for a biped walking pattern generation},
  author={Kajita, Shuuji and Kanehiro, Fumio and Kaneko, Kenji and Yokoi, Kazuhito and Hirukawa, Hirohisa},
  booktitle={IEEE/RSJ International Conference on Intelligent Robots and Systems (IROS)},
  volume={1},
  pages={239--246},
  year={2001}
}

@inproceedings{stasse2006real,
  title={Real-time 3d slam for humanoid robot considering pattern generator information},
  author={Stasse, Olivier and Davison, Andrew J and Sellaouti, Ramzi and Yokoi, Kazuhito},
  booktitle={IEEE/RSJ International Conference on Intelligent Robots and Systems (IROS)},
  pages={348--355},
  year={2006}
}

@inproceedings{gu2024conceptgraphs,
  title={Conceptgraphs: Open-vocabulary 3d scene graphs for perception and planning},
  author={Gu, Qiao and Kuwajerwala, Ali and Morin, Sacha and Jatavallabhula, Krishna Murthy and Sen, Bipasha and Agarwal, Aditya and Rivera, Corban and Paul, William and Ellis, Kirsty and Chellappa, Rama and others},
  booktitle={IEEE International Conference on Robotics and Automation (ICRA)},
  pages={5021--5028},
  year={2024}
}

@inproceedings{gaussian_grouping,
  title={Gaussian grouping: Segment and edit anything in 3d scenes},
  author={Ye, Mingqiao and Danelljan, Martin and Yu, Fisher and Ke, Lei},
  booktitle={European conference on computer vision (ECCV)},
  pages={162--179},
  year={2024},
  organization={Springer}
}

@article{bajcsy1988active,
  title={Active perception},
  author={Bajcsy, Ruzena},
  journal={Proceedings of the IEEE},
  volume={76},
  number={8},
  pages={966--1005},
  year={1988},
  publisher={IEEE}
}

@article{bohg2017interactive,
  title={Interactive perception: Leveraging action in perception and perception in action},
  author={Bohg, Jeannette and Hausman, Karol and Sankaran, Bharath and Brock, Oliver and Kragic, Danica and Schaal, Stefan and Sukhatme, Gaurav S},
  journal={IEEE Transactions on Robotics},
  volume={33},
  number={6},
  pages={1273--1291},
  year={2017}
}

@article{ho2020denoising,
  title={Denoising diffusion probabilistic models},
  author={Ho, Jonathan and Jain, Ajay and Abbeel, Pieter},
  journal={Advances in neural information processing systems (NeurIPS)},
  volume={33},
  pages={6840--6851},
  year={2020}
}

@article{hughes2022hydra,
  title={Hydra: A Real-time Spatial Perception Engine for 3D Scene Graph Construction and Optimization},
  author={Hughes, Nathan and Chang, Yun and Carlone, Luca},
  journal={Robotics: Science and Systems (RSS)},
  year={2022}
}

@inproceedings{kerr2023lerf,
  title={Lerf: Language embedded radiance fields},
  author={Kerr, Justin and Kim, Chung Min and Goldberg, Ken and Kanazawa, Angjoo and Tancik, Matthew},
  booktitle={Proceedings of the IEEE/CVF International Conference on Computer Vision (CVPR)},
  pages={19729--19739},
  year={2023}
}

@article{mildenhall2021nerf,
  title={Nerf: Representing scenes as neural radiance fields for view synthesis},
  author={Mildenhall, Ben and Srinivasan, Pratul P and Tancik, Matthew and Barron, Jonathan T and Ramamoorthi, Ravi and Ng, Ren},
  journal={Communications of the ACM},
  volume={65},
  number={1},
  pages={99--106},
  year={2021},
  publisher={ACM New York, NY, USA}
}

@article{muller2022instant,
  title={Instant neural graphics primitives with a multiresolution hash encoding},
  author={M{\"u}ller, Thomas and Evans, Alex and Schied, Christoph and Keller, Alexander},
  journal={ACM transactions on graphics},
  volume={41},
  number={4},
  pages={1--15},
  year={2022},
  publisher={ACM New York, NY, USA}
}

@inproceedings{qiu2024featuresplatting,
  title={Language-driven physics-based scene synthesis and editing via feature splatting},
  author={Qiu, Ri-Zhao and Yang, Ge and Zeng, Weijia and Wang, Xiaolong},
  booktitle={European conference on computer vision (ECCV)},
  pages={368--383},
  year={2024}
}

@inproceedings{radford2021learning,
  title={Learning transferable visual models from natural language supervision},
  author={Radford, Alec and Kim, Jong Wook and Hallacy, Chris and Ramesh, Aditya and Goh, Gabriel and Agarwal, Sandhini and Sastry, Girish and Askell, Amanda and Mishkin, Pamela and Clark, Jack and others},
  booktitle={International Conference on Machine Learning (ICML)},
  pages={8748--8763},
  year={2021}
}

@inproceedings{rana2023sayplan,
  title={SayPlan: Grounding Large Language Models using 3D Scene Graphs for Scalable Robot Task Planning},
  author={Rana, Krishan and Haviland, Jesse and Garg, Sourav and Abou-Chakra, Jad and Reid, Ian and Suenderhauf, Niko},
  booktitle={Conference on Robot Learning (CoRL)},
  pages={23--72},
  year={2023},
  organization={PMLR}
}

@inproceedings{chang2024goat,
  title={GOAT: GO to Any Thing},
  author={Chang, Matthew and Gervet, Th{\'e}ophile and Khanna, Mukul and Yenamandra, Sriram and Shah, Dhruv and Min, So Yeon and Shah, Kavit and Paxton, Chris and Gupta, Saurabh and Batra, Dhruv and others},
  booktitle={Robotics: Science and Systems (RSS)},
  year={2024}
}

@inproceedings{rosinol20203d,
  title={3D Dynamic Scene Graphs: Actionable Spatial Perception with Places, Objects, and Humans},
  author={Rosinol, Antoni and Gupta, Arjun and Abate, Marcus and Shi, Jingnan and Carlone, Luca},
  booktitle={Robotics: Science and Systems (RSS)},
  year={2020}
}

@inproceedings{yan2023gs,
  title={Gs-slam: Dense visual slam with 3d gaussian splatting},
  author={Yan, Chi and Qu, Delin and Xu, Dan and Zhao, Bin and Wang, Zhigang and Wang, Dong and Li, Xuelong},
  booktitle={Proceedings of the IEEE/CVF Conference on Computer Vision and Pattern Recognition (CVPR)},
  pages={19595--19604},
  year={2024}
}

@article{chen2025sam,
  title={Sam 3d: 3dfy anything in images},
  author={Chen, Xingyu and Chu, Fu-Jen and Gleize, Pierre and Liang, Kevin J and Sax, Alexander and Tang, Hao and Wang, Weiyao and Guo, Michelle and Hardin, Thibaut and Li, Xiang and others},
  journal={arXiv preprint arXiv:2511.16624},
  year={2025}
}

@inproceedings{siddiqui2024meshgpt,
  title={Meshgpt: Generating triangle meshes with decoder-only transformers},
  author={Siddiqui, Yawar and Alliegro, Antonio and Artemov, Alexey and Tommasi, Tatiana and Sirigatti, Daniele and Rosov, Vladislav and Dai, Angela and Nie{\ss}ner, Matthias},
  booktitle={Proceedings of the IEEE/CVF Conference on Computer Vision and Pattern Recognition (CVPR)},
  pages={19615--19625},
  year={2024}
}

@article{oquab2024dinov2,
  title={DINOv2: Learning Robust Visual Features without Supervision},
  author={Oquab, Maxime and Darcet, Timoth{\'e}e and Moutakanni, Th{\'e}o and Vo, Huy and Szafraniec, Marc and Khalidov, Vasil and Fernandez, Pierre and Haziza, Daniel and Massa, Francisco and El-Nouby, Alaaeldin and others},
  journal={Transactions on Machine Learning Research Journal},
  pages={1--31},
  year={2024}
}

@inproceedings{qin2024langsplat,
  title={Langsplat: 3d language gaussian splatting},
  author={Qin, Minghan and Li, Wanhua and Zhou, Jiawei and Wang, Haoqian and Pfister, Hanspeter},
  booktitle={Proceedings of the IEEE/CVF Conference on Computer Vision and Pattern Recognition (CVPR)},
  pages={20051--20060},
  year={2024}
}

@article{wilson2025latentbki,
  title={LatentBKI: Open-dictionary continuous mapping in visual-language latent spaces with quantifiable uncertainty},
  author={Wilson, Joey and Xu, Ruihan and Sun, Yile and Ewen, Parker and Zhu, Minghan and Barton, Kira and Ghaffari, Maani},
  journal={IEEE Robotics and Automation Letters},
  year={2025}
}

@article{oh2024deblurgs,
  title={Deblurgs: Gaussian splatting for camera motion blur},
  author={Oh, Jeongtaek and Chung, Jaeyoung and Lee, Dongwoo and Lee, Kyoung Mu},
  journal={arXiv preprint arXiv:2404.11358},
  year={2024}
}

@inproceedings{hanson2025pup,
  title={Pup 3d-gs: Principled uncertainty pruning for 3d gaussian splatting},
  author={Hanson, Alex and Tu, Allen and Singla, Vasu and Jayawardhana, Mayuka and Zwicker, Matthias and Goldstein, Tom},
  booktitle={Proceedings of the Computer Vision and Pattern Recognition Conference (CVPR)},
  pages={5949--5958},
  year={2025}
}

@article{schmid2024khronos,
  title={Khronos: A unified approach for spatio-temporal metric-semantic slam in dynamic environments},
  author={Schmid, Lukas and Abate, Marcus and Chang, Yun and Carlone, Luca},
  booktitle={Robotics: Science and Systems (RSS)},
  year={2024}
}

@inproceedings{quigley2009ros,
  title={ROS: an open-source Robot Operating System},
  author={Quigley, Morgan and Conley, Ken and Gerkey, Brian and Faust, Josh and Foote, Tully and Leibs, Jeremy and Wheeler, Rob and Ng, Andrew Y and others},
  booktitle={IEEE International Conference on Robotics and Automation Workshop on Open Source Software},
  volume={3},
  number={3.2},
  pages={5},
  year={2009}
}

\clearpage
\renewcommand{\baselinestretch}{1}
\setlength{\belowcaptionskip}{0pt}

\begin{strip}
\begin{center}
\vspace{-5ex}
\textbf{\LARGE \bf
Learning to Evolve: Multi-modal Interactive Fields for Robust Humanoid Navigation in Dynamic Environments
} \\
\vspace{3ex}

\Large{\bf- Supplementary Material -}\\
\vspace{0.6cm}

\vspace{0.5cm}
\hrule
\vspace{0.3cm}

\begin{minipage}{0.96\textwidth}
\normalsize
\textit{
The supplementary material provides implementation details, algorithmic workflow, parameter settings, and additional visual evidence for the Multi-modal Interactive Field (MIF) system reported in the main paper.
These details clarify how the appearance, spatial, geometry, navigation, and adaptation modules are integrated in the real humanoid deployment.
We include the distributed system architecture, execution loop, selected hyperparameters, and navigation-control details.
We also provide qualitative examples of locomotion-robust mapping, local memory updates, and IPS-based interaction checks.
}
\end{minipage}

\vspace{0.5cm}
\hrule
\vspace{0.4cm}

\noindent\textbf{\large Outline of Supplementary Material}

\vspace{0.2cm}
\begin{minipage}{0.96\textwidth}
\normalsize
This document provides additional implementation details, extended analyses, and qualitative examples that complement the main manuscript. The content is organized as follows:

\vspace{0.1cm}
\begin{itemize} \setlength{\itemsep}{0.4em} \setlength{\parskip}{0pt} \setlength{\parsep}{0pt}
    \item \textbf{\hyperref[sec:implementation_details]{Section S.1: System Implementation Details}}
    \begin{itemize} \setlength{\itemsep}{0.2em} \setlength{\topsep}{0.2em}
        \item \textbf{Hardware Architecture:} Specifications of the Unitree G1 humanoid platform, RealSense D435i sensor setup, and onboard/offboard computation split.
        \item \textbf{Software Stack:} Implementation details for the confidence-aware 3DGS backend, VLM-assisted scene-graph construction, and Flow Matching-based geometry service.
        \item \textbf{System Architecture:} A dataflow diagram showing asynchronous coordination among perception, spatial reasoning, geometry generation, and navigation.
    \end{itemize}

    \item \textbf{\hyperref[sec:algo_workflow]{Section S.2: Algorithmic Workflow}}
    \begin{itemize} \setlength{\itemsep}{0.2em} \setlength{\topsep}{0.2em}
        \item \textbf{Algorithm 1 (MIF Execution Loop):} Pseudo-code describing the closed-loop interaction between semantic grounding, stability-aware navigation, and the Memory Evolution Loop.
    \end{itemize}

    \item \textbf{\hyperref[sec:hyperparameters]{Section S.3: Hyperparameter Settings}}
    \begin{itemize} \setlength{\itemsep}{0.2em} \setlength{\topsep}{0.2em}
        \item \textbf{Confidence Gating Parameters:} Specific values for gradient penalty ($\beta$) and opacity scaling ($\gamma$) used to suppress locomotion-induced noise.
        \item \textbf{Interaction \& Adaptation:} Weighting coefficients ($w_{pos}, w_{sem}, w_{rel}$) for the multi-modal discrepancy score $\mathcal{D}$ and re-planning thresholds.
        \item \textbf{Safety Protocol:} Geometric margins ($\delta_{safe}$) and support polygon definitions for the Interaction Pose Safety (IPS) check.
    \end{itemize}

    \item \textbf{\hyperref[sec:add_experiments]{Section S.4: Additional Quantitative Results}}
    \begin{itemize} \setlength{\itemsep}{0.2em} \setlength{\topsep}{0.2em}
        \item \textbf{Ablation Study (Confidence Gating):} Extended PSNR/SSIM and change-detection measurements illustrating the effect of confidence gating under walking-induced distortion.
        \item \textbf{Navigation Stability Analysis:} Measurements of torso oscillation and cross-track error under the stability-aware local planner.
    \end{itemize}

    \item \textbf{\hyperref[sec:qualitative_results]{Section S.5: More Qualitative Visualizations}}
    \begin{itemize} \setlength{\itemsep}{0.2em} \setlength{\topsep}{0.2em}
        \item \textbf{Robustness of Appearance Field:} Visual examples comparing mapping with and without confidence gating, plus examples of local appearance updates after scene changes.
        \item \textbf{Geometric Safety Verification:} Comparison between sparse Gaussian centroids and generated meshes for IPS collision checking.
        \item \textbf{Dynamic Adaptation Sequence:} Chronological visualization of the Memory Evolution Loop in a ``Sofa Relocation'' (P2) scenario.
        \item \textbf{Real-World Deployment Showcase:} Examples of social navigation, narrow-passage traversal, dynamic adaptation, and mobile manipulation.
    \end{itemize}
\end{itemize}
\end{minipage}

\vspace{0.3cm}
\hrule
\vspace{0.8cm}

\end{center}
\end{strip}

\setcounter{section}{0}
\setcounter{equation}{0}
\setcounter{figure}{0}
\setcounter{table}{0}
\setcounter{page}{1}

\renewcommand{\thesection}{S.\arabic{section}}
\renewcommand{\thesubsection}{S.\arabic{section}-\Alph{subsection}}
\renewcommand{\thetable}{S.\arabic{table}}
\renewcommand{\thefigure}{S.\arabic{figure}}

\clearpage
\section{System Implementation Details}
\label{sec:implementation_details}

This section describes the hardware and software setup used in our real-world MIF experiments.

\subsection{Hardware Architecture}

\subsubsection{Robot Platform and Sensors}
We deploy our system on the Unitree G1 humanoid robot. The robot is equipped with a head-mounted Intel RealSense D435i RGB-D camera, configured to provide synchronized color and depth streams at a resolution of $640 \times 480$ (30 Hz). Camera intrinsics are calibrated offline, and the camera-to-base transform is obtained from the calibrated mounting and robot kinematic chain.

\subsubsection{Computation and Communication}
We split computation between onboard control and offboard perception/reasoning:

\begin{itemize}
    \item {Onboard Compute:} The robot's built-in NVIDIA Jetson Orin NX handles robot-provided low-level control interfaces, state feedback (IMU/Odometry), and hardware bridging.
    \item {Offboard Compute:} The high-load MIF modules, including 3DGS mapping, VLM-assisted graph construction, and Flow Matching-based geometry generation, run on the workstation.
    \item {Communication Stack:} We use ROS1 Noetic and compressed image transport for RGB-D streams. In our setup, this configuration provided approximately 25 Hz transmission with latency below 50 ms.
\end{itemize}

\subsubsection{Server Specifications}
The offboard workstation uses commercial desktop hardware. It is equipped with an Intel Core i9-13900K CPU, 64 GB DDR5 RAM, and an NVIDIA RTX 4090 GPU with 24 GB VRAM. This setup was sufficient for the reported 22 FPS Appearance Field updates and approximately 6.2 s Geometry Field generation latency in our experiments.

\subsection{Software Stack}

\subsubsection{3DGS Backend}

The Appearance Field is implemented on an incremental Gaussian Splatting backend. We use differentiable Gaussian rasterization libraries and CUDA kernels for primitive updates and rendering. The SLAM frontend is implemented in Python/PyTorch, leveraging {diff-gaussian-rasterization-w-depth} for differentiable rasterization with depth supervision. The confidence gate ($\Omega_i$) is implemented in PyTorch/CUDA and applied during rendering and pose optimization with low runtime overhead.

\subsubsection{VLM \& Prompt Engineering}

For Spatial Field ($\mathcal{F}_{spat}$) construction, we use {GPT-4o} (OpenAI) to parse stabilized rendered views into structured scene-graph evidence. The parsed outputs are post-processed into scene graph nodes and relations.

To encourage structured output for topological parsing, we use the JSON-style prompt template shown in Fig.~\ref{fig:prompt_engineering}. This prompt requests object categories, 2D bounding box estimates, and spatial prepositions (e.g., ``on'', ``next to'') relative to known landmarks.

\begin{figure}[h]
\centering
\begin{tcolorbox}[
    colback=gray!5,       
    colframe=black!70,    
    arc=3pt,              
    boxrule=0.6pt,        
    left=8pt, right=8pt, top=8pt, bottom=8pt, 
    width=0.95\columnwidth
]
\footnotesize
\renewcommand{\baselinestretch}{1.1}\selectfont 

\noindent \texttt{\textbf{System Instruction:}} \\
You are a robot's spatial reasoning engine. Analyze the provided image (synthesized from a stable view) and extract key objects.

\vspace{1.5ex}
\noindent \texttt{\textbf{Input:}} \\
An image of an office scene.

\vspace{1.5ex}
\noindent \texttt{\textbf{Output Constraint:}} \\
Return a JSON-style list. Each item should contain:
\vspace{0.5ex}
\begin{itemize} \setlength{\itemsep}{0pt} \setlength{\topsep}{0pt} \setlength{\parsep}{0pt}
    \item[\texttt{-}] \texttt{"object\_id": unique integer}
    \item[\texttt{-}] \texttt{"category": specific object name (e.g., "monitor", not "electronics")}
    \item[\texttt{-}] \texttt{"relations": list of spatial relations to other objects (e.g., ["on", "desk\_1"])}
    \item[\texttt{-}] \texttt{"attributes": visual descriptors (e.g., "black", "off")}
\end{itemize}

\vspace{1.5ex}
\noindent \texttt{\textbf{User Query:}} \\
"Identify all interactable objects in this view and their spatial relationships."

\end{tcolorbox}

\captionsetup{hypcap=false}
\caption{\textbf{Prompt template for scene graph construction.} The prompt requests structured JSON-style output that can be parsed into the topological graph $\mathcal{M}_{spat}$.}
\label{fig:prompt_engineering}
\end{figure}

\subsubsection{Flow Matching Model Architecture}

The Geometry Field ($\mathcal{F}_{geom}$) uses a pre-trained conditional Flow Matching model~\cite{chen2025sam}. Using a pre-trained model avoids training a geometry prior from scratch and provides object-level mesh hypotheses for IPS verification.

\begin{itemize}
    \item {Backbone Architecture:} We utilize a pre-trained Flow-Matching Transformer backbone that processes latent shape tokens and produces object-level mesh hypotheses used by the IPS module.
    
    \item {Conditioning \& Tokenization:} The conditioning image is encoded with frozen visual features and projected into the model's latent space, allowing the generation process to use both visual and geometric cues.
    
    \item {Inference Strategy:} We use a distilled inference setting with $N=25$ function evaluation steps. In our setup, geometry generation takes about 6.2 s and is scheduled asynchronously during approach or verification.
\end{itemize}

\subsubsection{System Architecture Implementation}
\label{sec:system_arch}

\begin{figure*}[h]
    \centering
    \includegraphics[width=\linewidth]{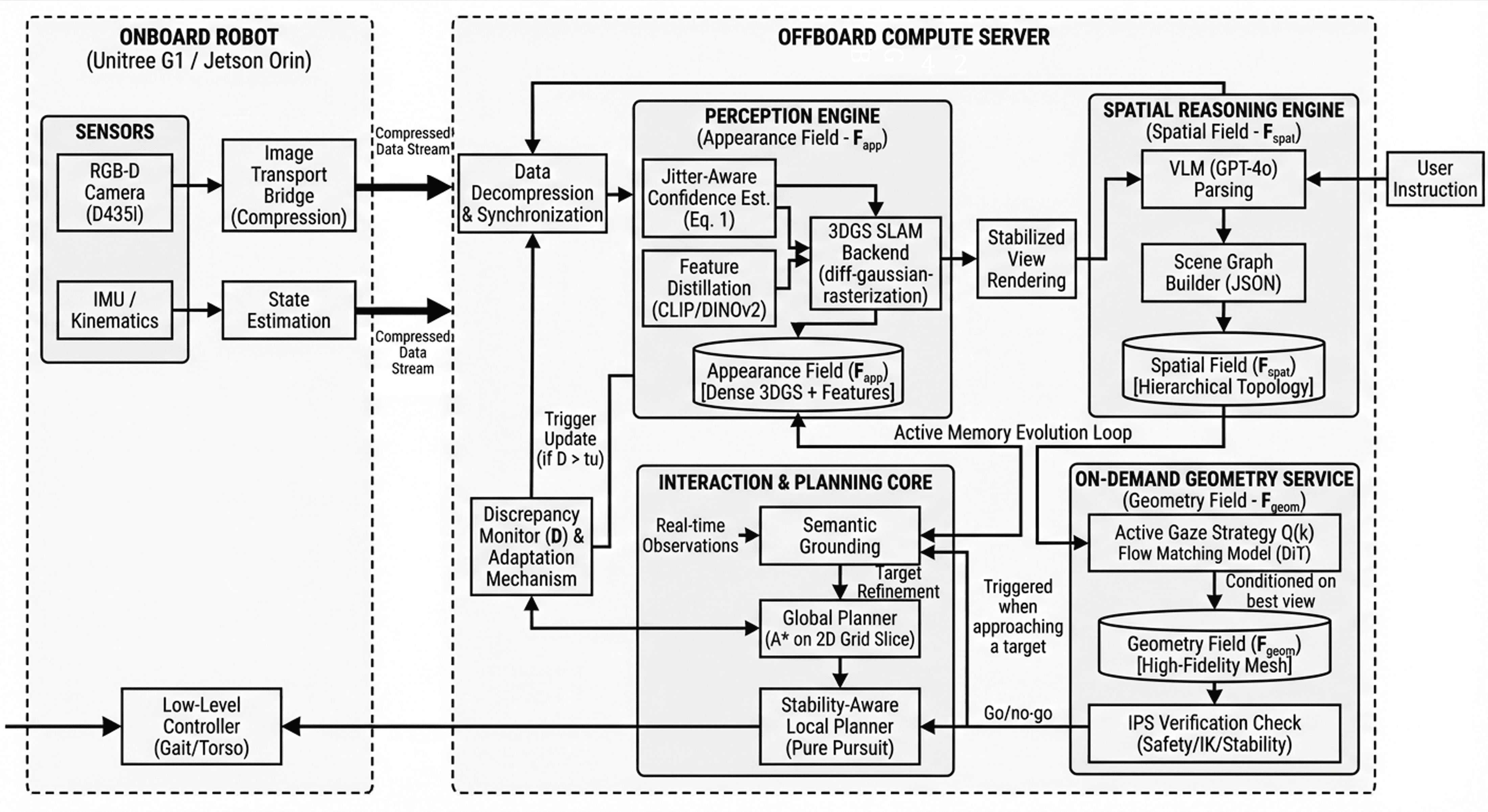}
    \caption{\textbf{System architecture and dataflow.} The architecture is divided between the {Onboard Robot} for sensing and low-level control, and the {Offboard Compute Server} for high-load perception and reasoning. On the server, processing is structured into four main engines: {Perception Engine} (constructing $\mathcal{F}_{app}$), {Spatial Reasoning Engine} (constructing $\mathcal{F}_{spat}$), {Interaction \& Planning Core}, and the {On-Demand Geometry Service} (generating $\mathcal{F}_{geom}$). The {Discrepancy Monitor} connects online observations with local memory updates when map-reality mismatch is detected.}
    \label{fig:sys_arch}
\end{figure*}

Fig.~\ref{fig:sys_arch} summarizes the onboard/offboard dataflow used by MIF.

The system comprises distinct computational modules distributed as follows:

\textbf{Onboard Computation (Unitree G1 with Jetson Orin):}
The onboard layer provides sensor streams, robot state feedback, and low-level command execution.
\begin{itemize}
    \item {Sensors:} Acquires raw data from the RGB-D Camera (D435i) and IMU/Kinematics.
    \item {Image Transport Bridge:} Compresses high-bandwidth RGB-D streams to minimize latency before transmission to the offboard server.
    \item {Low-Level Interface:} Executes admissible velocity and body commands through the robot control stack.
\end{itemize}

\textbf{Offboard Compute Server:}
Incoming compressed streams are first handled by the {Data Decompression \& Synchronization} module before being dispatched to four primary engines:

\begin{itemize}
    \item {Perception Engine (Appearance Field - $\mathcal{F}_{app}$):} Maintains the confidence-aware Appearance Field using the 3DGS SLAM backend, reliability estimation, and CLIP/DINOv2 feature distillation.

    \item {Spatial Reasoning Engine (Spatial Field - $\mathcal{F}_{spat}$):} Processes stabilized rendered views with VLM-assisted parsing and scene graph construction to build local graph evidence and update the Spatial Field when triggered by the adaptation loop.

    \item {Interaction \& Planning Core:} Uses real-time observations for semantic grounding and target refinement, runs the global planner and stability-aware local planner, and sends control commands back to the onboard robot.

    \item {On-Demand Geometry Service (Geometry Field - $\mathcal{F}_{geom}$):} Triggered when approaching an interaction target. It uses the task-local viewpoint score $Q(k)$ to select conditioning views for Flow Matching-based mesh generation, which is then checked by IPS.
\end{itemize}

The discrepancy monitor compares live local graph evidence with stored Spatial Field evidence. When $\mathcal{D}>\tau$, it triggers local updates in the Appearance and Spatial Fields rather than global remapping. Algorithm~\ref{alg:main_loop} summarizes this logic.

\section{Hyperparameter Settings}
\label{sec:hyperparameters}

We list the fixed parameter values used in our experiments. Unless otherwise noted, these values were kept unchanged across real-world trials.

\subsection{Confidence Gating Parameters}
In the Appearance Field ($\mathcal{F}_{app}$), the confidence score $C_i$ (Eq.~\ref{eq:confidence}) uses the sharpness of the gradient penalty ($\beta$) and the opacity scaling factor ($\gamma$). The values used in our experiments, along with the confidence threshold, are listed in Table~\ref{tab:supp_params_confidence}.

\begin{table}[h]
\centering
\footnotesize
\caption{\textbf{Parameters for confidence-aware filtering.}}
\label{tab:supp_params_confidence}
\begin{tabular*}{\linewidth}{@{\extracolsep{\fill}}ccl@{}}
\toprule
\textbf{Symbol} & \textbf{Value} & \textbf{Description} \\ \midrule
$\beta$ & 5.0 & Controls sensitivity to normalized optimization instability. \\
$\gamma$ & 2.0 & Scaling factor for opacity-based reliability. \\
$\tau_{conf}$ & 0.6 & Minimum confidence threshold for VLM rendering. \\ \bottomrule
\end{tabular*}
\end{table}

\subsection{Interaction \& Adaptation Mechanism}
The discrepancy score $\mathcal{D}$ (Eq.~\ref{eq:delta_calc}, \ref{eq:total_discrepancy}) aggregates positional, semantic, and relational graph evidence to trigger local memory update and replanning. The weighting coefficients for these components and the triggering threshold are listed in Table~\ref{tab:supp_params_discrepancy}.

\begin{table}[h]
\centering
\footnotesize
\caption{\textbf{Parameters for discrepancy detection.}}
\label{tab:supp_params_discrepancy}
\begin{tabular*}{\linewidth}{@{\extracolsep{\fill}}ccl@{}}
\toprule
\textbf{Symbol} & \textbf{Value} & \textbf{Description of Weight or Threshold} \\ \midrule
$w_{pos}$ & 1.0 & Centroid Euclidean distance error (in meters). \\
$w_{sem}$ & 0.5 & CLIP/DINO semantic feature cosine distance. \\
$w_{rel}$ & 0.8 & Jaccard distance of graph edge sets. \\
$\tau$ & 0.45 & Threshold of $\mathcal{D}$ for local memory update and replanning. \\ \bottomrule
\end{tabular*}
\end{table}

\subsection{Interaction Pose Safety (IPS) Protocol}
The IPS check is used as a pre-interaction feasibility gate.
\begin{itemize}
    \item {Geometric Safety Margin ($\delta_{safe}$):} We set $\delta_{safe} = 0.05\,\text{m}$. In our implementation, a pose is considered collision-free only if the minimum signed distance between the robot model and the generated mesh $\mathcal{M}_{gen}$ is greater than $5\,\text{cm}$.
    \item {Support Polygon ($\mathcal{P}_{sup}$):} Defined as the convex hull of the robot's feet contact area, with a safety shrinkage of $2\,\text{cm}$ to account for control noise.
\end{itemize}

\section{Algorithmic Workflow}
\label{sec:algo_workflow}

Algorithm~\ref{alg:main_loop} summarizes the task execution loop, including semantic grounding, navigation monitoring, local memory update, and IPS verification.

MIF is a system-level integration rather than a standalone 3DGS, scene-graph, or mesh-generation algorithm. The Appearance Field $\mathcal{F}_{app}$ stores confidence-gated semantic Gaussian primitives; the Spatial Field $\mathcal{F}_{spat}$ stores a hierarchical scene graph whose nodes contain semantic labels, centroids, reliability, and room affiliation; and the Geometry Field $\mathcal{F}_{geom}$ is an on-demand object mesh used only for IPS verification. The gradient magnitude in $\mathcal{F}_{app}$ is not used as a generic 3DGS densification or pruning heuristic; here it is interpreted as a physical instability cue caused by humanoid gait and pose jitter. During adaptation, MIF only revises the graph nodes, edges, and Gaussian support inside the discrepancy-triggered local region, while unaffected memory remains unchanged.

\begin{algorithm*}[t]
\caption{Multi-modal Interactive Field (MIF) execution loop}
\label{alg:main_loop}
\begin{algorithmic}[1]
\REQUIRE User Instruction $I$, Initial Memory $\mathcal{M} = \{\mathcal{F}_{app}, \mathcal{F}_{spat}\}$
\STATE \textbf{Phase 1: Semantic Grounding (Sec. 3.4)}
\STATE $\langle \text{Region}, \text{Target} \rangle \leftarrow \text{VLM\_Parse}(I)$
\STATE $\mathbf{v}_{target} \leftarrow \text{QueryGraph}(\mathcal{F}_{spat}, \text{Target})$ \COMMENT{Retrieve target node}
\STATE $\mathbf{c}_{goal} \leftarrow \text{RefineCentroid}(\mathbf{v}_{target}, \mathcal{F}_{app})$ \COMMENT{Eq. 3: Confidence-weighted position}
\STATE $\mathcal{P}_{global} \leftarrow \text{PlanPath}(\mathbf{c}_{goal})$

\STATE \textbf{Phase 2: Navigation \& Monitoring (Sec. 3.2)}
\WHILE{distance($\mathbf{p}_{robot}, \mathbf{c}_{goal}$) $> \delta_{arrival}$}
    \STATE $\mathcal{O}_{loc} \leftarrow \text{GetObservation()}$
    \STATE $\mathcal{M}_{loc} \leftarrow \text{BuildLocalGraph}(\mathcal{O}_{loc})$ \COMMENT{Live nodes, edges, and centroids}
    \STATE $\mathcal{D} \leftarrow \text{CalcDiscrepancy}(\mathcal{M}_{loc}, \mathcal{F}_{spat})$ \COMMENT{Eq. 5: Monitor P2 Mismatch}
    
    \IF{$\mathcal{D} > \tau$}
        \STATE \textbf{Trigger Memory Evolution Loop:}
        \STATE Pause navigation.
        \STATE $\mathcal{O}_{scan} \leftarrow \text{ActiveScan}(\text{Region})$ \COMMENT{Gather multi-view evidence}
        \STATE $\mathcal{R}_{bad} \leftarrow \text{AffectedRegion}(\mathcal{M}_{loc}, \mathcal{M}_{spat})$ \COMMENT{Nodes/edges with high discrepancy}
        \STATE $\mathcal{F}_{app}[\mathcal{R}_{bad}] \leftarrow \text{FuseHighConf}(\mathcal{O}_{scan})$ \COMMENT{Local Gaussian support only}
        \STATE $\mathcal{M}_{loc}^{new} \leftarrow \text{BuildLocalGraph}(\mathcal{F}_{app}[\mathcal{R}_{bad}])$
        \STATE $\mathcal{F}_{spat} \leftarrow \text{PatchGraph}(\mathcal{F}_{spat}, \mathcal{M}_{loc}^{new})$ \COMMENT{Replace/insert affected nodes and edges}
        \STATE $\mathbf{c}_{goal} \leftarrow \text{ReQuery}(\mathcal{F}_{spat})$ \COMMENT{Refresh target estimate}
        \STATE $\mathcal{P}_{global} \leftarrow \text{PlanPath}(\mathbf{c}_{goal})$ \COMMENT{Re-plan on updated map}
    \ELSE
        \STATE $v, \omega \leftarrow \text{StabilityAwareControl}(\mathcal{P}_{global})$ \COMMENT{Sec. S.1.3}
        \STATE $\text{Execute}(v, \omega)$
    \ENDIF
\ENDWHILE

\STATE \textbf{Phase 3: Interaction Verification (Sec. 3.3)}
\WHILE{$S_{IPS} == \text{False}$ and retry budget remains}
    \STATE $k^* \leftarrow \text{SelectViewpoint}(Q(k))$ \COMMENT{Eq. 6: Active Gaze}
    \STATE $\text{MoveTo}(k^*)$
    \STATE $I_{best} \leftarrow \text{CaptureImage}()$
    \STATE $\mathcal{M}_{gen} \leftarrow \text{FlowMatching}(I_{best}, \text{cond}=\mathbf{v}_{target}.\text{sem})$ \COMMENT{Eq. 7}
    \STATE $S_{IPS} \leftarrow \text{CheckIPS}(\mathcal{M}_{gen})$ \COMMENT{Eq. 9: Safety Check}
    
    \IF{$S_{IPS} == \text{False}$}
        \STATE $\text{MicroAdjustStance}(\mathcal{M}_{gen})$ \COMMENT{Adjust stance or select another viewpoint}
    \ENDIF
\ENDWHILE

\RETURN \text{Task status}
\end{algorithmic}
\end{algorithm*}

\section{Additional Quantitative Results}
\label{sec:add_experiments}

This section provides extended quantitative analyses that complement the task-level results in the main paper. We report an ablation of confidence gating under walking-induced distortion and a stability analysis of the local planner.

\subsection{Ablation Study: Efficacy of Confidence Gating}
\label{sec:ablation_study}

To isolate the effect of confidence gating ($\Omega_i$) defined in Eq.~\ref{eq:confidence}, we compare MIF with a variant without the gate (``w/o Gating'').

{Quantitative Analysis.} We evaluate the rendered-view quality (PSNR and SSIM) under different robot walking speeds. As shown in Table~\ref{tab:supp_ablation_gating}, the ungated variant degrades under fast walking ($0.5\,\text{m/s}$), with PSNR dropping to 21.6 dB. MIF retains higher PSNR/SSIM under the same condition. The lower false-positive rate (FPR) further suggests that stable appearance evidence helps reduce false updates in the adaptation loop.

\begin{table}[h]
\centering
\footnotesize
\caption{\textbf{Extended ablation of confidence gating under walking motion.}}
\label{tab:supp_ablation_gating}
\begin{tabular*}{\linewidth}{@{\extracolsep{\fill}}lccccc@{}}
\toprule
\multirow{2}{*}{\textbf{Method}} & \multicolumn{2}{c}{\textbf{Slow Walk ($0.2\,\text{m/s}$)}} & \multicolumn{2}{c}{\textbf{Fast Walk ($0.5\,\text{m/s}$)}} & \textbf{Change Det.} \\ \cmidrule(lr){2-3} \cmidrule(lr){4-5} \cmidrule(lr){6-6}
 & PSNR $\uparrow$ & SSIM $\uparrow$ & PSNR $\uparrow$ & SSIM $\uparrow$ & FPR $\downarrow$ \\ \midrule
w/o Gating & 28.4 & 0.88 & 21.6 & 0.72 & 46.5\% \\
\textbf{MIF (Ours)} & \textbf{31.2} & \textbf{0.93} & \textbf{29.8} & \textbf{0.89} & \textbf{4.2\%} \\ \bottomrule
\end{tabular*}
\end{table}

\subsection{Navigation Stability Analysis}
We also report torso oscillation and cross-track error for the stability-aware local planner. We define \textit{Torso Oscillation} as the standard deviation of the camera's roll/pitch angles during traversal. 
Compared to a standard Pure Pursuit controller, adaptive velocity scaling reduces torso oscillation by {45\%} (from $5.2^\circ$ to $2.8^\circ$) while maintaining a low cross-track error ($<0.15\,\text{m}$). This reduction in oscillation is consistent with the improved perceptual quality reported in Table~\ref{tab:supp_ablation_gating}.

\section{Qualitative Visualizations}
\label{sec:qualitative_results}

This section provides qualitative examples illustrating the system's behavior under humanoid locomotion noise (P1) and dynamic scene changes (P2).

\subsection{Robust Mapping under Locomotion (P1)}

\subsubsection{Robustness of Appearance Field: Jitter Suppression \& Dynamic Adaptation}

\begin{figure*}[t]
    \centering
    \includegraphics[width=\linewidth]{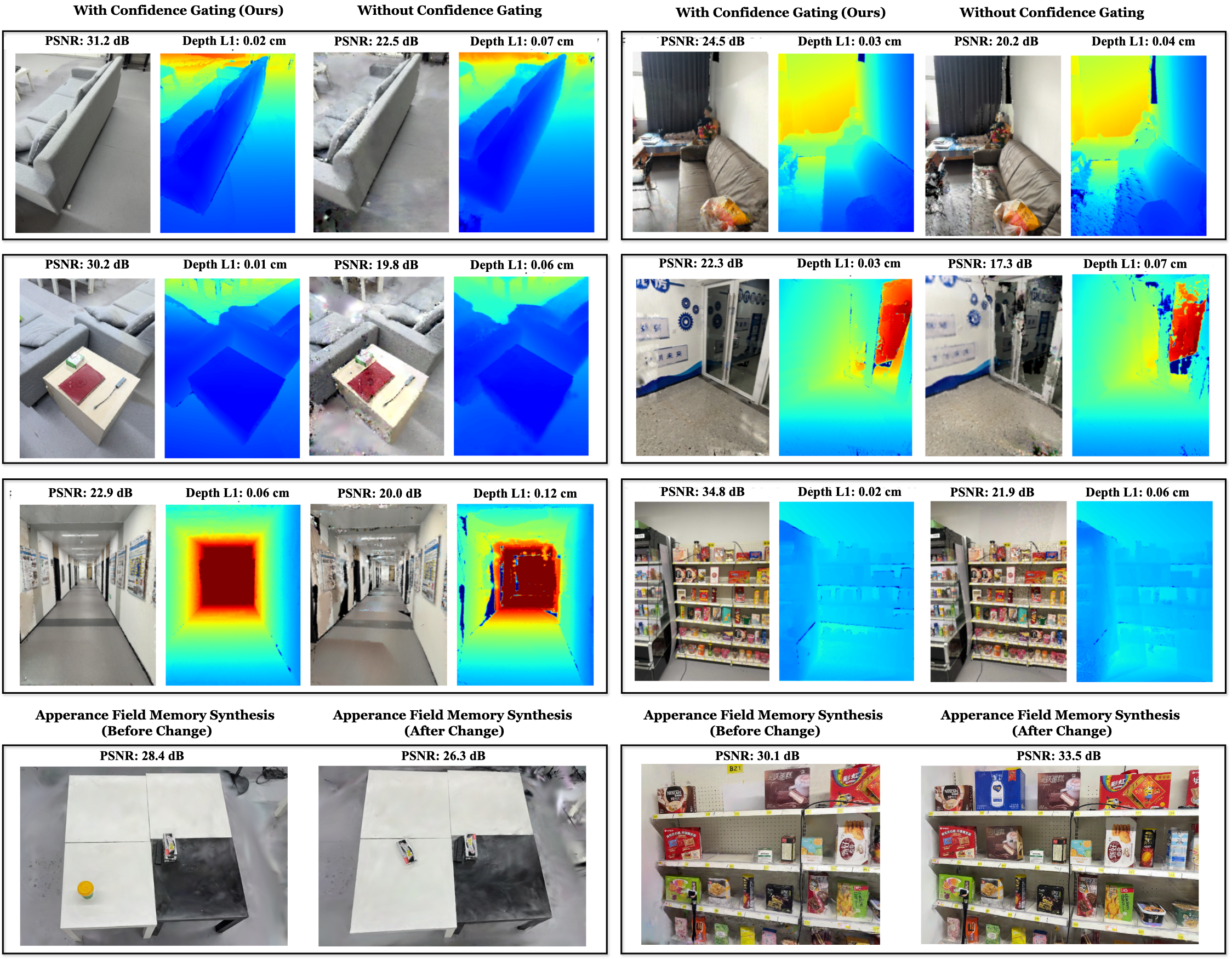}
    \caption{\textbf{Qualitative Evaluation of Appearance Field Robustness.} 
    \textbf{(Rows 1-3) Ablation of Confidence Gating:} We compare rendered RGB and depth maps between our method (Left) and the baseline without gating (Right) under locomotion noise. \textbf{Without gating}, the map shows motion blur (ghosting) and depth discontinuities. \textbf{With confidence gating}, the system reduces visible jitter artifacts, yielding higher PSNR and lower Depth L1 errors.
    \textbf{(Row 4) Dynamic Adaptation:} The bottom row visualizes an Appearance Field update after environmental changes (e.g., object relocation on the table/shelf) while maintaining visual consistency in the rendered views.}
    \label{fig:supp_jitter}
\end{figure*}

We visualize rendered RGB/depth quality from the Appearance Field $\mathcal{F}_{app}$ under humanoid walking. {Rows 1-3 of Fig.~\ref{fig:supp_jitter}} illustrate the effect of confidence gating during high-speed walking ($0.5\,\text{m/s}$). The ungated variant shows ghosting artifacts in RGB and noisier depth estimates (e.g., Depth L1 error increases to $0.07\,\text{cm}$). In contrast, confidence gating prioritizes higher-reliability observations, maintaining high PSNR ($>30\,\text{dB}$) and piecewise-smooth geometry in this example.

Furthermore, {Row 4} shows an example in which local field updates reflect object removal or placement without visibly accumulating motion artifacts.

\subsubsection{Generated Geometry for IPS Checking}

\begin{figure*}[t]
    \centering
    \includegraphics[width=0.95\linewidth]{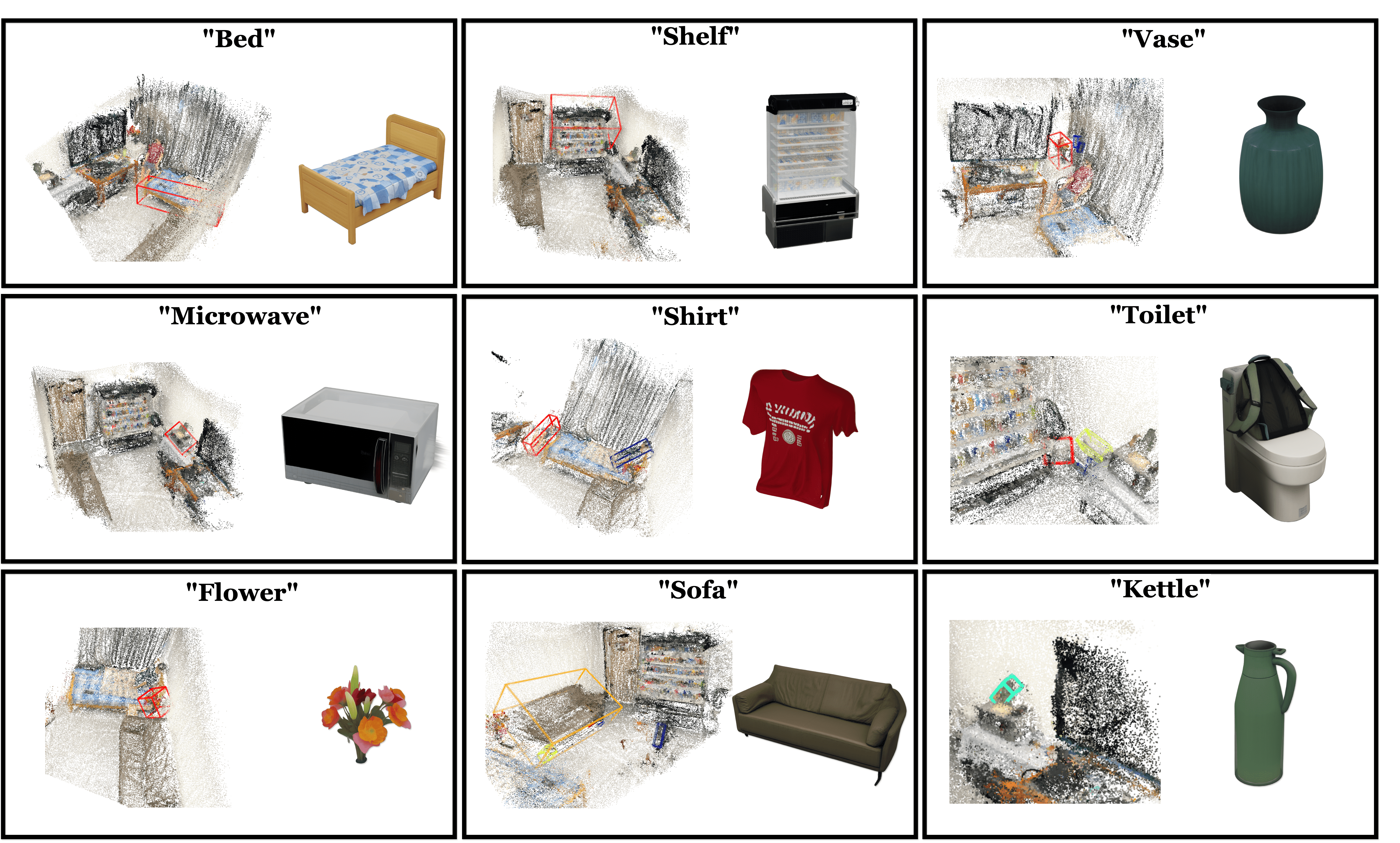}
    \caption{\textbf{From sparse perception to interaction checking.} Comparison between the raw environmental point cloud and the generated Geometry Field. \textbf{(Left)} The sparse 3D Gaussian centroids capture the scene context but contain noise and gaps (visualized with bounding boxes), making them insufficient for fine-grained collision checks. \textbf{(Right)} The corresponding meshes generated by $\mathcal{F}_{geom}$ provide a denser geometric proxy for collision and reachability checks.}
    \label{fig:pclvsmesh}
    
    \vspace{0.2cm}

    \includegraphics[width=0.95\linewidth]{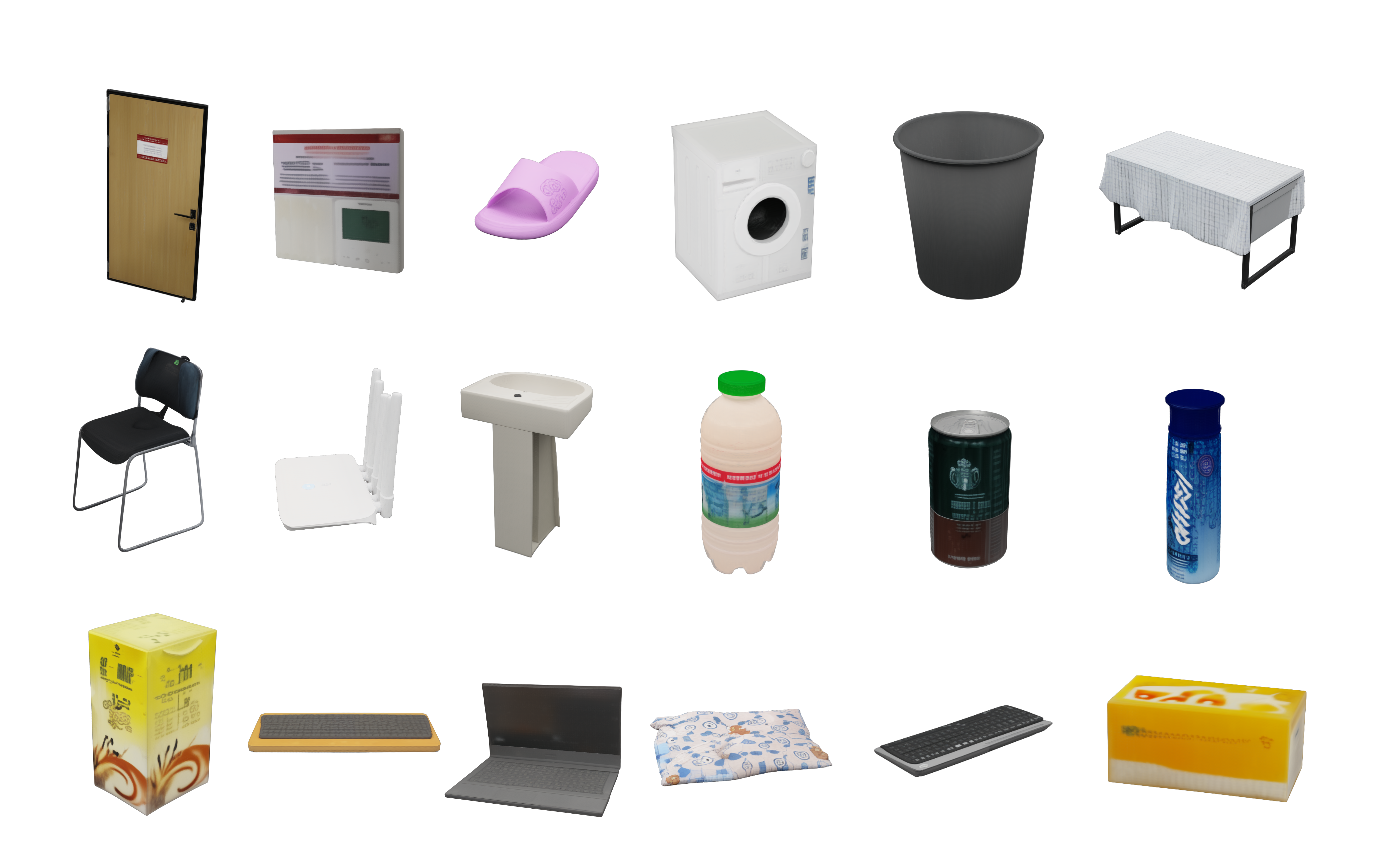}
    \caption{\textbf{Gallery of generated interactive geometries.} Examples of meshes generated by the Flow Matching model during real-world experiments, covering rigid appliances (e.g., washing machines, routers), furniture (e.g., chairs, tables), and semi-rigid objects (e.g., pillows). These assets are used as geometric proxies for the Interaction Pose Safety (IPS) check.}
    \label{fig:meshes}
\end{figure*}

To illustrate why sparse point clouds are insufficient for IPS collision checking, we present a comparative analysis in Fig.~\ref{fig:pclvsmesh}. The raw 3D Gaussian point cloud (Left panels), while useful for visual navigation, remains sparse and lacks surface connectivity. Gaps between centroids may make collision checks overly optimistic.

The Flow Matching module synthesizes a dense mesh (Right panels) for the target object. The generated mesh provides a continuous proxy for signed-distance checks in IPS. Fig.~\ref{fig:meshes} further shows examples across several object categories used in our experiments.

\subsection{Dynamic Adaptation Sequence (P2)}
\begin{figure}[H]
    \centering
    \includegraphics[width=\linewidth]{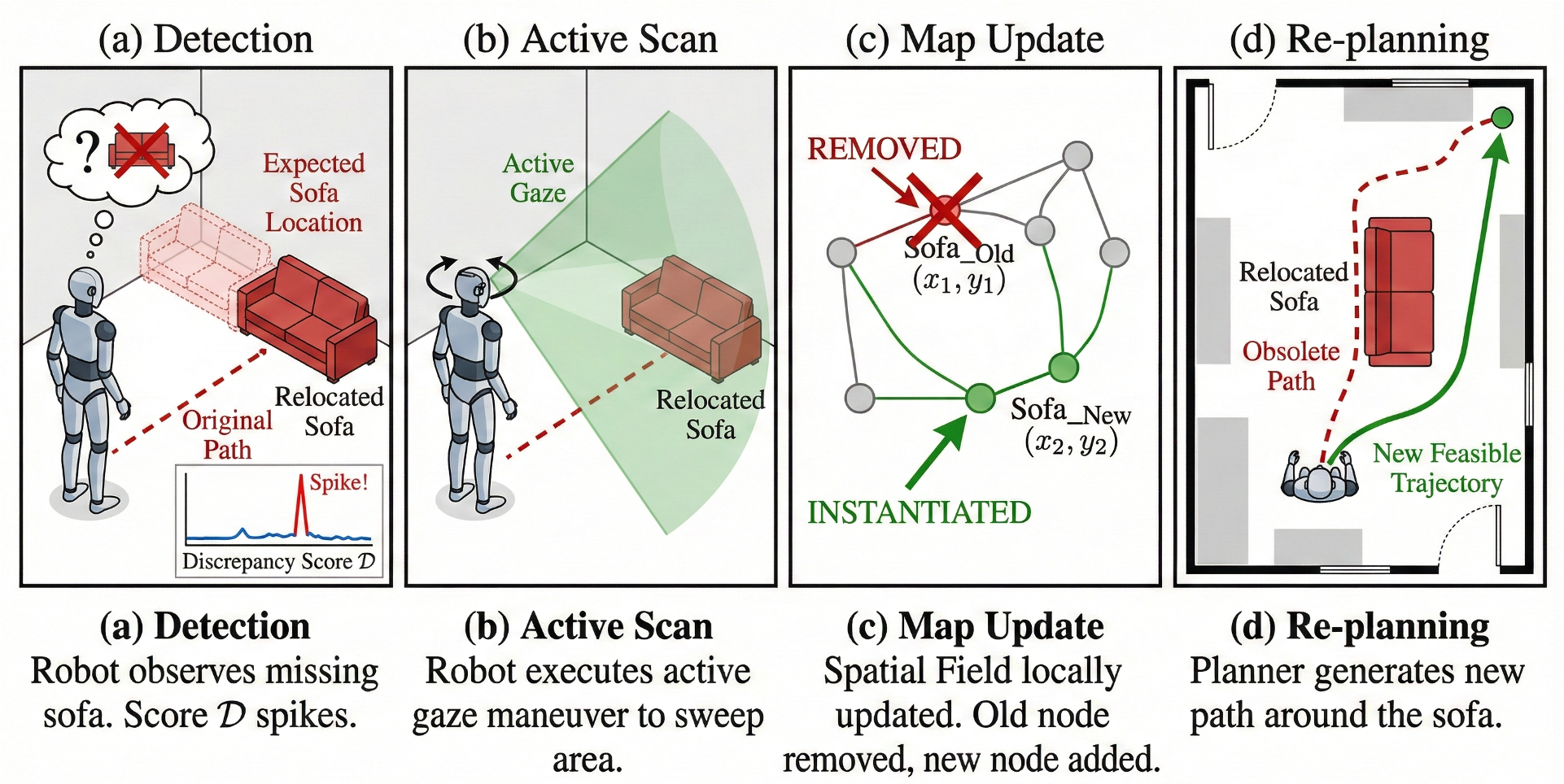}
    \caption{\textbf{Memory Evolution Loop under sofa relocation.}
    \textbf{(a)} The discrepancy score $\mathcal{D}$ rises when the sofa is missing.
    \textbf{(b)} The robot actively scans the area.
    \textbf{(c)} The local graph is updated with the observed sofa location.
    \textbf{(d)} The planner re-routes to the updated goal.}
    \label{fig:supp_sequence}
\end{figure}

Fig.~\ref{fig:supp_sequence} presents a step-by-step breakdown of the Interaction Adaptation Mechanism in a real-world "Sofa Relocation" scenario.

\subsection{Real-World Deployment Showcase}
\label{sec:real_world_showcase}

We provide additional deployment examples on the Unitree G1 humanoid. As shown in Fig.~\ref{fig:showoff}, these examples cover social navigation, narrow-passage traversal, dynamic adaptation, and a mobile-manipulation task.

\noindent\textbf{Social Navigation (Row 1).} The robot patrols a corridor, detects a human, stops at a socially compliant distance ($1.5\,\text{m}$), and executes a greeting gesture before resuming navigation.

\noindent\textbf{Cluttered Navigation (Row 2).} The robot traverses a narrow office passage toward the target area, illustrating the navigation stack operating with the maintained Appearance Field $\mathcal{F}_{app}$.

\noindent\textbf{Dynamic Adaptation (Rows 3-5).} In the "Sofa Relocation" scenario (P2), the baseline (HOV-SG, Row 5) follows stale scene-graph memory toward the obsolete sofa location, while MIF (Row 4) detects a discrepancy, updates local memory, and replans toward the observed layout.

\noindent\textbf{Mobile Manipulation (Row 6).} Finally, we show a service task in an elderly care setting. The robot navigates to a fridge, retrieves a drink, and hands it to a user. This example illustrates how the generated Geometry Field $\mathcal{F}_{geom}$ can support IPS-based interaction checks before manipulation.

\begin{figure*}[t]
    \centering
    \includegraphics[width=0.9\linewidth]{images/showoff.jpg}
    \caption{\textbf{Real-world deployment examples.} We show: 
\textbf{Row 1:} Social patrol with compliant human interaction (waving at $1.5\,\text{m}$). 
\textbf{Row 2:} Narrow-passage traversal through an office passageway. 
\textbf{Rows 3-5:} Dynamic adaptation to layout changes (``\textit{Bypass the sofa}''). The baseline follows stale memory (Row 5), while MIF updates local memory and replans (Row 4). 
\textbf{Row 6:} Mobile-manipulation example for fetching a drink.}
    \label{fig:showoff}
\end{figure*}

\end{document}